\documentclass[3p,a4paper]{elsarticle}
\usepackage[utf8]{inputenc}
\usepackage{subfigure}

\usepackage{makecell}
\title{MINPO: Memory-Informed Neural Pseudo-Operator to Resolve Nonlocal Spatiotemporal Dynamics  
}

\author[myUaddress]{Farinaz Mostajeran}
\author[myUaddress]{Aruzhan Tleubek}
\author[myUaddress]{Salah A Faroughi \corref{mycorrespondingauthor}}

\cortext[mycorrespondingauthor]{Corresponding author: salah.faroughi@utah.edu}

\address[myUaddress]{Energy \& Intelligence Lab, Department of Chemical Engineering, University of Utah, Salt Lake City, Utah  84112, USA
}

\date{\today}

\usepackage{mwe}

\usepackage{setspace} 
\usepackage{tabularx}
\usepackage{lineno,hyperref}
\usepackage{amsmath}
\usepackage{bm}
\usepackage{amssymb}
\usepackage[]{amsmath}
\usepackage{upgreek}
\usepackage{float}
\usepackage{booktabs, multirow, makecell}

\let\today\relax
\makeatletter
\def\ps@pprintTitle{%
    \let\@oddhead\@empty
    \let\@evenhead\@empty
    \def\@oddfoot{\footnotesize\itshape
         {Submitted preprint — December 2025} \hfill\today}%
    \let\@evenfoot\@oddfoot
    }
\makeatother

\usepackage{pgfplots}
\pgfplotsset{compat=1.5}
\usepgfplotslibrary{units}
\usetikzlibrary{spy}
\usetikzlibrary{pgfplots.groupplots}

\usepackage{epstopdf}
\usepackage{units}
\usepackage{lscape}
\usepackage{booktabs}
\usepackage{gensymb}
\usepackage{multirow}

\usepackage{graphicx}
\usepackage{caption}
\usepackage{subcaption}

\usepackage{booktabs}
\usepackage{multirow}
\usepackage{caption}
\usepackage{subcaption}

\usepackage{soul,color}
\soulregister\cite7
\soulregister\ref7
\soulregister\pageref7

\usepackage[flushleft]{threeparttable}
\usepackage{tikz}
\usepackage{placeins}
\usepackage{color}
\usepackage{hyperref}

\usepackage{stackengine}

\usepackage{array}
\usepackage{tabularx}
\usepackage{pdfpages}
\usepackage{natbib}
\biboptions{numbers,sort,compress} 
\usepackage{textcomp, gensymb}
\usepackage{booktabs} 

\newtheorem{theorem}{Theorem}

\newtheorem{lemma}[theorem]{Lemma}

\newtheorem{open problem}[theorem]{Open Problem}
{\par\noindent\textbf{Proof.}\ }%
{\hfill$\square$\par}

\begin{document}

\begin{abstract}

Many physical systems exhibit nonlocal spatiotemporal behaviors described by integro-differential equations (IDEs). Classical methods for solving IDEs require repeatedly evaluating convolution integrals, whose cost increases quickly with kernel complexity and dimensionality. Existing neural solvers can accelerate selected instances of these computations, yet they do not generalize across diverse nonlocal structures. In this work, we introduce the Memory-Informed Neural Pseudo-Operator (MINPO), a unified framework for modeling nonlocal dynamics arising from long-range spatial interactions and/or long-term temporal memory. MINPO, employing either Kolmogorov-Arnold Networks (KANs) or multilayer perceptron networks (MLPs) as encoders, learns the nonlocal operator and its inverse directly through neural representations, and then explicitly reconstruct the unknown solution fields. The learning is guarded by a lightweight nonlocal consistency loss term to enforce coherence between the learned operator and reconstructed solution. The MINPO formulation allows to naturally capture and efficiently resolve nonlocal spatiotemporal dependencies governed by a wide spectrum of IDEs and their subsets, including fractional PDEs. We evaluate the efficacy of MINPO in comparison with classical techniques and state-of-the-art neural-based strategies based on MLPs, such as A-PINN and fPINN, along with their newly-developed KAN variants, A-PIKAN and fPIKAN, designed to facilitate a fair comparison.  Our study offers compelling evidence of the accuracy of MINPO and demonstrates its robustness in handling (i) diverse kernel types, (ii) different kernel dimensionalities, and (iii) the substantial computational demands arising from repeated evaluations of kernel integrals. MINPO, thus, generalizes beyond problem-specific formulations, providing a unified framework for systems governed by nonlocal operators.

\end{abstract}


\maketitle

\section{Introduction}\label{sec:Intro}

Integro-differential equations (IDEs) describe systems whose evolution depends not only on local, instantaneous states but also on information accumulated over space and time. This gives rise to nonlocal spatiotemporal dynamics, encompassing both spatial nonlocality arising from long-range spatial interactions and temporal memory associated with long-term temporal dynamics~\cite{belair1989consumer, yu2016fractional, pang2019fpinns, yuan2022pinn, failla2020advanced, nikan2020numerical, amirali2025second}. Such behaviors appear across many applications, including viscoelastic materials~\cite{yu2016fractional, mainardi2022fractional}, biological and ecological systems~\cite{wang2022fractional, ma2023bi, zinihi2025identifying}, anomalous diffusion~\cite{wang2023physics,epps1803turbulence,ma2025adaptive}, and Euler–Bernoulli beam~\cite{wang2016exact,romano2017constitutive}, and structural dynamics~\cite{mostajeran2025solving,marwah2023neural,minakov2021integro}.
A major difficulty is that many IDEs, especially fractional-order models, contain power-law kernels that induce long-term temporal dynamics. Evaluating the solution thus often requires retaining and updating its full time history, which results in high memory usage and significant computational cost~\cite{dahlby2011general,sorvari2010time}.
More broadly, numerical approximation of IDEs is challenged by three factors: (i) the diversity and possible singularity of memory kernels~\cite{belkina2016dynamical,yuldashev2021new}, (ii) the growth in the number and dimensionality of integral terms~\cite{mahdy2023computational,pachpatte2011multidimensional}, and (iii) the substantial effort required to repeatedly evaluate these operators~\cite{chu2009direct,michaels2012parallel}. To address these issues, methods have evolved from classical semi-analytical and discretization schemes~\cite{gao2014new, zhao2017numerical, sayyar2021high} to modern neural-network–based formulations~\cite{ lu2021deepxde, guo2022monte, wang2024gmc}.

Most classical semi-analytical methods for solving IDEs are inherently ad-hoc, requiring the user to reformulate the original IDE into a form compatible with the chosen technique. The most prominent examples, Adomian decomposition~\cite{adomian2013solving, adomian1994decomposition, duan2012review}, differential transforms~\cite{arikoglu2005solution, darania2007method, arikoglu2008solutions}, and perturbation methods\cite{biazar2009he, roul2011numerical, omotosho2020modified, rahmah2025novel}, offer elegant representations, but require extensive symbolic manipulation and remain highly problem-specific, limiting their applicability across a wider range of IDEs. A large body of purely numerical techniques has also been developed for IDEs, mainly finite difference, but also spectral, finite element and finite volume methods for specific cases. Spectral methods approximate the solution with global polynomial bases (e.g., Chebyshev, Legendre, Laguerre, or fractional variants) and rewrite the IDE as a system of algebraic equations for the expansion coefficients~\cite{kajani2006numerical, hosseini2003tau, talaei2022numerical, fakhar2011spectral}. Similar to semi-analytical approaches, they require the user to adapt the problem to the method and to select several method-dependent parameters such as the basis order, number of collocation nodes, and domain partitioning. Finite difference schemes are typically designed for narrowly defined subclasses of IDEs, depending on the equation’s structure~\cite{mbroh2020second, cakir2022numerical,raji2025computational} (Volterra or Fredholm), nonlinearity, type of nonlocality~\cite{guo2020finite, luo2022numerical}(time-memory or spatial), and kernel regularity~\cite{tang1993finite, xu2020compact,mohebbi2017compact}, and therefore do not generalize well to more complex IDEs. Even high-order finite difference methods apply only to equations with smooth kernels and struggle when nonlinearities or kernel singularities are present~\cite{zhao2006compact}. Finite volume methods are inherently tailored to local conservation laws~\cite{eymard2000finite, barth2003finite}, whereas IDEs introduce nonlocal interactions across the entire domain. As a result, finite volume methods are generally suitable only under strong structural assumptions, such as in specialized studies of variable-coefficient parabolic IDEs~\cite{gan2020efficient}. Finite element formulations face similar limitations, often relying on stringent regularity assumptions or special quadrature rules tailored to the kernel structure~\cite{li2010finite, chen2019two, yi2015h}. Moreover, because the integral term couples each point to its entire history, all these methods lead to iterative solvers whose computational cost grows rapidly with problem size. Altogether, the existing IDE literature lacks a unified numerical framework that is cost-effective and applicable across general nonlinear and singular-kernel IDEs.

Recently, neural-network–driven solvers have gained significant attention for addressing aforementioned challenges faced by classical numerical methods in resolving IDEs, as they can be formulated as general-purpose, mesh-independent, and tolerance-agnostic learning frameworks capable of accommodating a wide range of IDE classes. Specifically, advances in physics-informed neural networks (PINNs)~\cite{raissi2019physics, faroughi2024physics} have shown that these models can be extended beyond standard differential equations to handle integro-differential formulations~\cite{ye2024fbsjnn, bassi2024learning}. 
These extensions augment the standard PINN architecture with mechanisms for representing nonlocal integral operators, either through numerical approximations or learned auxiliary fields, offering a pathway to treat the nonlocal spatiotemporal dynamics characteristic of IDEs.
Early physics-informed neural solvers approximate the memory operator through numerical quadrature inside the training loop, for example, DeepXDE~\cite{lu2021deepxde} employs Gauss-Legendre rules for Volterra integrals, which reduces mesh dependence but still requires an external discretization of the kernel. More advanced strategies, such as the Auxiliary PINN (A-PINN) framework proposed by Yuan et al.~\cite{yuan2022pinn}, further minimize quadrature by learning auxiliary fields that represent the nonlocal integral operator itself, enabling a fully mesh-free reconstruction of temporal memory terms. While these neural formulations effectively mitigate the challenges associated with discretization and growing dimensionality, they remain inherently limited by the structure of the nonlocal operator they can encode. In particular, models designed around Volterra-type or low-dimensional kernels often fail to capture spatial nonlocality, temporal memory, or their combination as nonlocal spatiotemporal dynamics, leading to noticeable accuracy loss.

Fractional-order IDEs introduce challenges that extend beyond those in standard integro-differential models, motivating the development of hybrid neural–numerical strategies that combine neural networks with classical discretization techniques~\cite{ma2023pmnn, sm2024novel}. Their defining operators, such as the Caputo~\cite{kai2004analysis} and Riemann–Liouville formulations~\cite{samko1993fractional, lischke2018fractional}, involve power-law kernels that generate long-range temporal interactions and inherently depend on the full history of the solution.
To resolve this, Fractional PINNs (fPINNs)~\cite{pang2019fpinns} were introduced that integrate automatic differentiation with discrete Gr\"{u}nwald-Letnikov approximations of fractional operators~\cite{zhao2015series, lischke2018fractional}. While effective, this coupling introduces discretization error and requires repeatedly evaluating long segments of the temporal history, a cost that grows rapidly with problem dimensionality. Monte-Carlo PINNs (MC-PINNs)~\cite{guo2022monte} and general Monte-Carlo PINNs (GMC-PINNS)~\cite{wang2024gmc} mitigate part of this overhead by replacing deterministic quadrature with stochastic sampling, while still relying on explicit operator approximations constructed through numerical sampling.
High-order finite-difference schemes, such as the method in~\cite{mostajeran2025solving}, combine an L-type approximation of temporal fractional derivatives~\cite{gao2014new, zhao2017numerical, sayyar2021high} with a step-wise learning strategy to improve accuracy through sharper representations of fractional dynamics. Even with higher-order formulas, these methods must still store and update large portions of the temporal history, and their cost grows with the number of time steps. Thus, they improve accuracy but do not resolve the core challenge posed by long-memory effects. Although such hybrid approaches expand the range of kernels and dimensions that neural solvers can handle, they remain tied to explicit discretization of the fractional operator. This reliance on fixed quadrature rules or numerical surrogates reduces flexibility, complicates adaptation to diverse kernel behaviors, and limits their ability to represent fractional and nonlocal operators in a truly model-based manner. Designing architectures that capture fractional dynamics and nonlocal spatiotemporal interactions without rigid discretization templates therefore remains an open challenge.

In this work, we introduce the Memory-Informed Neural Pseudo-Operator (MINPO), a unified framework designed to overcome the  limitations of existing neural and hybrid IDE solvers. The key innovation lies in learning a continuous neural representation of the memory operator itself, rather than discretizing it or constraining it to a particular kernel form. Through this representation, MINPO naturally accommodates a broad class of operators, including Volterra-type, spatial nonlocality arising from long-range spatial interactions, memory associated with long-term temporal dynamics, and fully nonlocal spatiotemporal dynamics, all within a single formulation. MINPO retains a continuous loss for the governing IDE while introducing a discretized nonlocal consistency loss that enforces agreement between the learned operator and the corresponding integral.
To demonstrate its generality, MINPO is implemented with both multilayer perceptrons (MLPs) and the more expressive Kolmogorov–Arnold Networks (KANs)~\cite{liu2024kan, liu2024kan2, bozorgasl2405wav, ss2024chebyshev}. 
While MLPs learn only weights and biases with fixed activations, KANs learn the activation functions themselves using adaptive basis components~\cite{yu2024kan, zeng2024kan}.
KANs have also begun to appear in scientific machine learning pipelines for solving differential equations~\cite{koenig2024kan, mostajeran2024epi}, most notably in PIKAN and DeepOKAN-type formulations~\cite{shukla2024comprehensive, xiapikans, mostajeran2025scaled,  abueidda2025deepokan, kiyani2025predicting}, demonstrating their potential as expressive and parameter-efficient alternatives to classical MLP architectures~\cite{jacob2024spikans,  faroughi2025neural, faroughi2025scientific, noorizadegan2025practitioner}. 
To demonstrate the benefits of MINPO, we perform a suite of benchmark experiments against well-established baselines, including A-PINN and fPINN. In addition, we construct their KAN-based counterparts (A-PIKAN and fPIKAN) to enable a fair and rigorous comparison of MINPO’s accuracy, computational efficiency, and ability to approximate operators exhibiting spatial nonlocality, memory effects, and nonlocal spatiotemporal dynamics. These comparisons collectively validate that MINPO offers a more flexible and broadly applicable approach to modeling complex IDEs.

The remainder of the paper is structured as follows.
Section~\ref{sec:minpo} introduces the MINPO framework and details its formulation.
Section~\ref{sec:results} presents a range of numerical experiments demonstrating accuracy, efficiency, and robustness across diverse kernel types and dimensions.
Section~\ref{Sec.Conclusion} summarizes the main findings and outlines potential directions for future work. In addition, we present several appendices that provide supporting material: ~\ref{app:preliminary} reviews the PINN, PIKAN, and scaled formulations used as baselines; ~\ref{A-PIKANN} and~\ref{fPIKANN} introduce the A-PIKAN and fPIKAN architectures introduced for comparison; and ~\ref{app:inv_M} provides a sample derivation illustrating how the unknown field can be reconstructed using the memory operator and its derivative for a representative class of kernels.


\section{MINPO Formulation }\label{sec:minpo}

We denote the spatiotemporal coordinate by $\boldsymbol{\xi}=(\boldsymbol{x},t)\in\Omega\times(0,T)=:\Omega_{\xi}$ and define its causal neighborhood by $\Omega_{\boldsymbol{\xi}}^{-}:=\{\,\boldsymbol{\eta}=(\boldsymbol{y},\tau)\in\Omega\times(0,t):\tau\le t\,\}$, representing all past space–time points that may influence the solution $u$ at $\boldsymbol{\xi}$. A general integro–differential equation (IDE) for systems exhibiting both spatial and temporal nonlocality is written in the unified form,
\begin{equation}
\mathcal{T}_\alpha[u](\boldsymbol{\xi})
=
\mathcal{N}[u](\boldsymbol{\xi})
+
\mathcal{M}[u](\boldsymbol{\xi})
+
S(\boldsymbol{\xi}),
\label{eq:general-IDE}
\end{equation}
where $\mathcal{N}[u](\boldsymbol{\xi})$ denotes the local nonlinear response of the field $u$ evaluated pointwise at $\boldsymbol{\xi}$, $S(\boldsymbol{\xi})$ is an external source term, and the temporal evolution operator is given by,
\begin{equation}
\mathcal{T}_\alpha[u](\boldsymbol{\xi})
=
\lambda_1\,\partial_t u(\boldsymbol{\xi})
+
\lambda_\alpha\,{}^{C}D_t^\alpha u(\boldsymbol{\xi}),
\qquad 0<\alpha<1,
\label{eq:T-operator}
\end{equation}
with ${}^{C}D_t^\alpha$ denoting the Caputo fractional derivative. The coefficient $\lambda_1$ controls the classical time derivative and $\lambda_\alpha$ governs the strength of fractional dynamics, which inherently encode temporal memory. All remaining nonlocal contributions (arising from spatial convolution kernels, spatial fractional operators, Volterra-type temporal memory, or mixed space–time interactions) are absorbed into a unified (possibly nonlinear) memory operator of the form,
\begin{equation}
\mathcal{M}[u](\boldsymbol{\xi})
:=
\int_{\Omega_{\boldsymbol{\xi}}^{-}}
K\!\bigl(\boldsymbol{\xi},\boldsymbol{\eta};u(\boldsymbol{\eta})\bigr)\,
\mathrm{d}\boldsymbol{\eta},
\label{eq:general-memory}
\end{equation}
where $K$ is a general interaction kernel describing how strongly the solution $u$ at $\boldsymbol{\xi}$ is influenced by its past values or by values of $u$ at other spatial locations $\boldsymbol{\eta}$ in space–time. Temporal
memory corresponds to $\boldsymbol{\eta}=(\boldsymbol{x},\tau)$ with $\tau\le t$ (i.e., the value of $u$ at time $t$ depends on all previous values of $u$ at that same location $\boldsymbol{x}$),
while spatial nonlocality corresponds to $\boldsymbol{\eta}=(\boldsymbol{y},t)$
with $\boldsymbol{y}\in\mathbb{R}^d$ (i.e., the value of $u$ at ($\boldsymbol{x},t$) depends on values of $u$ at other locations 
$\boldsymbol{y}$ at the same time).  Mixed space–time nonlocal interactions
are also naturally included in the same form.  Special cases follow directly: setting $\lambda_\alpha=0$ yields a classical IDE,
\begin{equation}
\partial_t u(\boldsymbol{\xi})
=
\mathcal{N}[u](\boldsymbol{\xi})
+
\mathcal{M}[u](\boldsymbol{\xi})
+
S(\boldsymbol{\xi}),
\end{equation}
while setting $\lambda_1=0$ produces a fractional IDE,
\begin{equation}
{}^{C}D_t^\alpha u(\boldsymbol{\xi})
=
\mathcal{N}[u](\boldsymbol{\xi})
+
\mathcal{M}[u](\boldsymbol{\xi})
+
S(\boldsymbol{\xi}),
\end{equation}
and choosing $\mathcal{M}[u]\equiv 0$ within the latter recovers a pure fractional PDE whose temporal nonlocality is encoded entirely by the Caputo operator  that is defined as, 
\begin{equation}\label{Eq:CaputoDeriv_1}
{}^{C}\!D_t^{\alpha} u(\boldsymbol{\xi})
=
\frac{1}{\Gamma(1-\alpha)}
\int_{0}^{t}
\frac{\partial_{t}u(\boldsymbol{x},\tau)}{(t-\tau)^{\alpha}}\,d\tau,
\end{equation}
    where $\Gamma(\cdot)$ is the Gamma function~\cite{kilbas2006theory}.

\begin{figure}[t]
    \centering
    \includegraphics[width=0.99\linewidth]{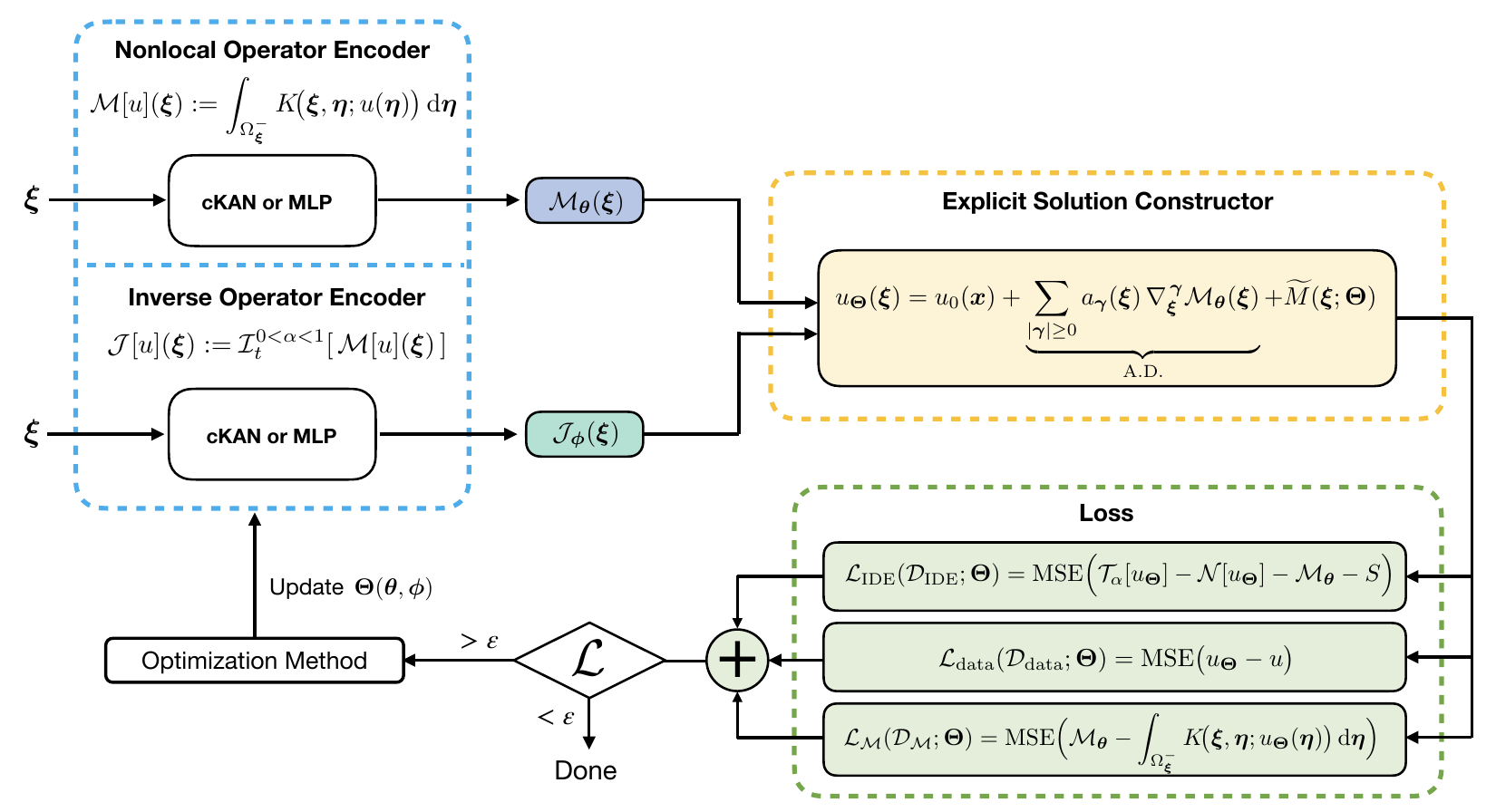}
    \caption{A schematic architecture for the proposed memory-informed neural pseudo-operator (MINPO) method to resolve IDEs with long-range spatial (nonlocal) interactions and/or long-term temporal (memory) dependencies.
    The network takes the spatiotemporal input coordinates $\boldsymbol{\xi}$ and outputs the learned  nonlocal operator $\mathcal{M}_{\boldsymbol{\theta}}(\boldsymbol{\xi})$, which provides a continuous neural representation of all spatiotemporal nonlocal effects. MINPO then  reconstructs the solution $u_{\boldsymbol{\Theta}}$ through an explicit map that incorporates $\mathcal{M}_{\boldsymbol{\theta}}$, its derivatives computed using automatic differentiation (A.D.), and (only for fractional models with $0<\alpha<1$) its inverse nonlocal operator component learned using a different network in parallel, $\mathcal{J}_{\boldsymbol{\phi}}(\boldsymbol{\xi})$.
    The training loss consists of three terms: a fully continuous IDE residual enforcing the governing physics, a data term enforcing initial/boundary/measurement constraints, and finally a lightweight nonlocal consistency term, $\mathcal{L}_{\mathcal{M}}$,  that compares the network output with the numerically evaluated nonlocal operator. MINPO can use various neural encoders, such as MLPs or KANs, yielding a versatile and expressive unified framework for solving IDEs.}
    \label{fig:MINPO-figure}
\end{figure}

In our MINPO framework, shown in Fig.~\ref{fig:MINPO-figure}, the memory operator $\mathcal{M}[u]$ is
parameterized by a neural network with $\boldsymbol{\theta}$ (implemented using a KAN or MLP encoders  as introduced in ~\ref{app:preliminary}),
\begin{equation}\label{class_def_network}
    \mathcal{M}_{\boldsymbol{\theta}}(\boldsymbol{\xi}) \approx \mathcal{M}[u](\boldsymbol{\xi}),
\end{equation}
and thus all required derivatives, spatial, temporal, or mixed, can be evaluated through automatic differentiation denoted as,
$
    \nabla_{\boldsymbol{\xi}}
    \mathcal{M}_{\boldsymbol{\theta}}(\boldsymbol{\xi}).
$
For fractional models, the unified memory representation must also capture the intrinsic temporal nonlocality introduced by the Caputo derivative. In this case, the memory network approximates the fractional derivative, while a second network represents its fractional inverse. Accordingly, the two learned fields satisfy the coupled relations,
\begin{equation}\label{Eq:frac_def_network}
    \mathcal{M}_{\boldsymbol{\theta}}(\boldsymbol{\xi}) \approx {}^{C}\!D_t^{\alpha} u(\boldsymbol{\xi}) ,
\qquad\text{and}\qquad
\mathcal{J}_{\boldsymbol{\phi}}(\boldsymbol{\xi}) \approx
\mathcal{I}_t^{\alpha}\!\bigl[\,{}^{C}\!D_t^{\alpha} u\,\bigr](\boldsymbol{\xi}) ,
\end{equation}
where \(\mathcal{I}_t^\alpha\) denotes the Riemann-Liouville fractional integral that serves as the inverse operator of the Caputo derivative parameterized by a second neural network with parameters \(\boldsymbol{\phi}\).
Thus, fractional models require two neural representations: one for the fractional memory field and one for its inverse fractional integral, as presented in Fig.~\ref{fig:MINPO-figure}.

Now, to recover the solution we introduce a general inverse time-integral operator, 
\begin{equation}
    \mathcal{I}_{t}^\alpha[f](t)
    :=
    \begin{cases}
        \displaystyle \int_0^t f(\tau)\,\mathrm{d}\tau,
        & \alpha = 1, \\[1.0em]
        \displaystyle
        \frac{1}{\Gamma(\alpha)}
        \int_0^t (t-\tau)^{\alpha-1} f(\tau)\,\mathrm{d}\tau,
        & 0<\alpha<1,
    \end{cases}
\end{equation}
so that $\mathcal{I}_{t}^1$ corresponds to the classical integral, and
$\mathcal{I}_{t}^\alpha$ inverts the Caputo derivative. Applying the inverse-time operator to the IDE yields the integral equation, 
\begin{equation}
    u(\boldsymbol{\xi})
    =
    u_0(\boldsymbol{x})
    +
     \mathcal{I}_{t}^\alpha
    \left[
        \mathcal{N}[u](\cdot)
        +
        \mathcal{M}[u](\cdot)
        +
        S(\cdot)
    \right](\boldsymbol{\xi}), 
\label{eq:unified-reconstruction}
\end{equation}
where  $u(\boldsymbol{\xi})$ denotes the unknown field evaluated at the
spatiotemporal coordinate $\boldsymbol{\xi}$, and
$u_0(\boldsymbol{x})$ is the prescribed initial condition at $t=0$.  The function $\mathcal{N}[u](\boldsymbol{\xi})$ collects all purely local
contributions (standard PDE-type terms depending on $u$ and its local
derivatives at $\boldsymbol{\xi}$). Equation~\eqref{eq:unified-reconstruction},  along with the definitions given in Eq.~\eqref{Eq:frac_def_network}, provides a  mathematical framework for reconstructing the solution in systems governed by purely
classical integro–differential equations, fractional PDEs, or mixed spatial–temporal nonlocal dynamics. However, this relation is
implicit in~$u$ because the local operator $\mathcal{N}[u](\boldsymbol{\xi})$
contains $u$ inside the time integral.  Consequently, this representation
cannot be used directly as an explicit pointwise reconstruction map
$u(\boldsymbol{\xi}) \leftarrow \mathcal{M}(\boldsymbol{\xi})$ in our proposed neural
pipeline. In special cases, e.g., Volterra-type equations, where the memory $\mathcal{M}$ is defined by a temporal
convolution of $u$, the Leibniz' rule generates a closed algebraic relation
between $u$ and $\mathcal{M_{\boldsymbol{\theta}}}$ and its finite set of temporal derivatives of $\mathcal{M}_{\boldsymbol{\theta}}$
(e.g.\ $u=\mathcal{M}_{\boldsymbol{\theta}}'+\mathcal{M}_{\boldsymbol{\theta}}$). Thus, the implicit general reconstruction reduces to an explicit expression for $u$ in terms of $\mathcal{M}$ and its temporal
derivatives, and all occurrences of $\mathcal{N}$ and $S$ become absorbed into the coefficients generated by the Leibniz identity.  Motivated by the Volterra case's closed algebraic relation  and Eq.\eqref{eq:unified-reconstruction}, we introduce an explicit reconstruction ansatz as,
\begin{equation}
u_{\boldsymbol{\Theta}}(\boldsymbol{\xi})
=
u_0(\boldsymbol{x})
+
\sum_{|\boldsymbol{\gamma}|\ge 0}
a_{\boldsymbol{\gamma}}(\boldsymbol{\xi})\,
\nabla_{\boldsymbol{\xi}}^{\,\boldsymbol{\gamma}}
\mathcal{M}_{\boldsymbol{\theta}}(\boldsymbol{\xi})
+
\widetilde{M}(\boldsymbol{\xi};\boldsymbol{\Theta}), \qquad \boldsymbol{\Theta} = (\boldsymbol{\phi}, \boldsymbol{\theta}),
\label{eq:recon-ansatz}
\end{equation}
that expresses the solution explicitly in terms of the learned memory field
$\mathcal{M}_{\boldsymbol{\theta}}$, its spatiotemporal derivatives, and learned inverse-memory contribution (needed only in fractional IDE cases; see the inverse memory encoder in Fig.~\ref{fig:MINPO-figure}). Here, $\nabla_{\boldsymbol{\xi}}^{\,\boldsymbol{\gamma}} \mathcal{M}_{\boldsymbol{\theta}}$
denotes a mixed spatiotemporal derivative of order
$|\boldsymbol{\gamma}|=\gamma_1+\cdots+\gamma_{d+1}$,
\begin{equation}
    \nabla_{\boldsymbol{\xi}}^{\,\boldsymbol{\gamma}}
    \mathcal{M}_{\boldsymbol{\theta}}(\boldsymbol{\xi})
    :=
    \frac{\partial^{|\boldsymbol{\gamma}|}
          \mathcal{M}_{\boldsymbol{\theta}}}
         {\partial x_1^{\gamma_1}\cdots
          \partial x_d^{\gamma_d}\,
          \partial t^{\gamma_{d+1}}}
    (\boldsymbol{\xi}),
\end{equation}
where $\boldsymbol{\gamma}=(\gamma_1,\ldots,\gamma_d,\gamma_{d+1})$ is a
multi-index on space and time.  The coefficient functions $a_{\boldsymbol{\gamma}}(\boldsymbol{\xi})$ are fixed analytic
coefficients obtained by applying the inverse-time operator
$\mathcal{I}^{\alpha}$ to the governing IDE and expanding the resulting memory
terms using Leibniz' rules. These coefficients
describe how derivatives of the unified memory field enter the closed-form
reconstruction of $u$ and are therefore not trainable parameters. The summation over $|\boldsymbol{\gamma}|\ge 0$ allows the IDE to involve the
memory field itself (for $|\boldsymbol{\gamma}|=0$) as well as its higher-order
spatiotemporal derivatives. Also, $\widetilde{M}(\boldsymbol{\xi}; \boldsymbol{\Theta})$ consists of temporal traces the memory operator in classical IDEs (e.g.,
$\mathcal{M}_{\boldsymbol{\theta}}(\boldsymbol{x},t=0)$) and both temporal traces  and fractional inverse-kernel contributions (i.e., $\mathcal{J}_{\boldsymbol{\phi}}(\boldsymbol{\xi})$) for fractional cases.
This ansatz thus provides an explicit, differentiable map
from the learned memory representation to the reconstructed solution, ensuring
that $u_{\boldsymbol{\Theta}}(\boldsymbol{\xi})$ can be obtained pointwise once
$\mathcal{M}_{\boldsymbol{\theta}}$ and $\mathcal{J}_{\boldsymbol{\phi}}$ (for
fractional models) are trained.

To train the MINPO model, we formulate a composite loss function that incorporates the physics residual, data constraints (when available),  and the consistency of the learned memory representation.  Let
\(\boldsymbol{\Theta} = (\boldsymbol{\theta},\boldsymbol{\phi})\) denote the full set of trainable parameters.
The total loss is written as (see Fig.~\ref{fig:MINPO-figure}),
\begin{equation}\label{Eq:LossMinpo}
\mathcal{L}(\boldsymbol{\Theta}) =
\lambda_{\text{IDE}}\, \mathcal{L}_{\text{IDE}}(\mathcal{D}_{\text{IDE}}; \boldsymbol{\Theta})
+
\lambda_{\text{data}}\, \mathcal{L}_{\text{data}}(\mathcal{D}_{\text{data}}; \boldsymbol{\Theta})
+
\lambda_{\mathcal{M}}\, \mathcal{L}_{\mathcal{M}}(\mathcal{D}_{\mathcal{M}}; \boldsymbol{\Theta}),
\end{equation}
where 
the coefficients $\lambda_{\text{IDE}}$, $\lambda_{\text{data}}$, and $\lambda_{\mathcal{M}}$ balance the contributions of the physics residual, data-fidelity terms, and memory consistency, respectively, and
each term serves a distinct role.
In Eq.~\eqref{Eq:LossMinpo}, the term \(\mathcal{L}_{\text{IDE}}\) measures how well the neural solution satisfies the governing integro–differential equation. The loss is defined through the mean-squared error (MSE) of the residual,
\begin{equation}
    \mathcal{L}_{\text{IDE}}(\mathcal{D}_{\text{IDE}}; \boldsymbol{\Theta})
=\text{MSE}\Big(
\mathcal{T}_\alpha[u_{\boldsymbol{\Theta}}] -
\mathcal{N}[u_{\boldsymbol{\Theta}}] -
\mathcal{M}_{\boldsymbol{\theta}} -
S
\Big),
\end{equation}
which penalizes the mismatch between the neural reconstruction and the governing equation, ensuring that the learned solution honors the underlying physics.
This residual is evaluated at a set of interior collocation points
\(\mathcal{D}_{\text{IDE}}=\{\boldsymbol{\xi}^{\text{IDE}}_i,\, i=1, \ldots, N_{\text{IDE}}\}\subset\Omega_{\xi}\).
At these points, the learned fields \(u_{\boldsymbol{\Theta}}\) and \(\mathcal{M}_{\boldsymbol{\theta}}\) are inserted into the network-based approximation of the IDE operator to compute the residual used in \(\mathcal{L}_{\text{IDE}}\).
A key distinction of MINPO is that the IDE loss \(\mathcal{L}_{\text{IDE}}\) is kept in fully continuous form. Unlike fPINNs~\cite{pang2019fpinns}, MINPO does not discretize the governing integro–differential equation inside the physics residual, and unlike A-PINNs~\cite{yuan2022pinn}, it does not introduce auxiliary outputs to represent intermediate integral terms. By avoiding such discretization of the governing equation, MINPO preserves the continuous structure of the IDE, allowing the residual to be evaluated at any resolution and at any interior point of the domain. This design leads to higher accuracy, eliminates the dependency on fixed quadrature grids, and allows MINPO to remain flexible with respect to different kernel types, dimensions, and memory structures.
In Eq.~\eqref{Eq:LossMinpo}, the data loss \(\mathcal{L}_{\text{data}}\) enforces agreement between the neural prediction and all available pointwise information. It is defined as,
\begin{equation}
    \mathcal{L}_{\text{data}} (\mathcal{D}_{\text{data}}; \boldsymbol{\Theta})
=
\text{MSE}\big(
u_{\boldsymbol{\Theta}} - u
\big),
\end{equation}
which ensures fidelity to physical or observational constraints and ties the learned solution to the available data.
The data employed in this loss are collected in
$
\mathcal{D}_{\text{data}}
=
\mathcal{D}_{\text{init}}
\cup
\mathcal{D}_{\text{bc}}
\cup
\mathcal{D}_{\text{meas}}
$
including initial conditions, boundary conditions, and, when provided, interior measurement data for both forward and inverse problems. The discrepancy between the predicted solution and these data points forms the basis of \(\mathcal{L}_{\text{data}}\).

The third loss component in Eq.~\eqref{Eq:LossMinpo}, denoted by \(\mathcal{L}_{\mathcal{M}}\), plays a central role in ensuring that the learned representation \(\mathcal{M}_{\boldsymbol{\theta}}\) truly behaves as the underlying memory operator of the governing IDE. This term explicitly enforces consistency between the network output and the nonlocal operator defined in Eqs.~\eqref{eq:general-memory} and \eqref{Eq:CaputoDeriv_1}, depending on whether the system exhibits general spatiotemporal nonlocality or intrinsic fractional temporal dynamics. Accordingly, the nonlocal consistency loss is defined through a mean-squared error that compares \(\mathcal{M}_{\boldsymbol{\theta}}\) with the corresponding continuous memory integral evaluated using the current neural estimate of the solution field \(u_{\boldsymbol{\Theta}}\). For IDEs governed by a general kernel \(K\), this loss reads,
\begin{equation}\label{Eq:Mloss_1}
    \mathcal{L}_{\mathcal{M}}(\mathcal{D}_{\mathcal{M}};\boldsymbol{\Theta})
=
\mathrm{MSE}\Big(
\mathcal{M}_{\boldsymbol{\theta}}
-
\int_{\Omega_{\boldsymbol{\xi}}^{-}}
K\!\bigl(\boldsymbol{\xi},\boldsymbol{\eta}; u_{\boldsymbol{\Theta}}(\boldsymbol{\eta})\bigr)\,
\mathrm{d}\boldsymbol{\eta}
\Big).
\end{equation}

In practice, evaluating these continuous integrals requires numerical quadrature, and we employ standard techniques such as Gauss–Legendre rules~\cite{swarztrauber2003computing, laurie2001computation} for general kernels or L-type discretizations for fractional operators~\cite{gao2014new, zhao2017numerical, sayyar2021high}.
The dataset \(\mathcal{D}_{\mathcal{M}}\subset\Omega_{\xi}\) is built by using the appropriate quadrature nodes along the integral dimension(s) and random sampling elsewhere, providing exactly the points at which the learned memory field is constrained to satisfy its integral definition.
In contrast to the continuous treatment of the IDE residual, the nonlocal consistency loss \(\mathcal{L}_{\mathcal{M}}\) requires numerical quadrature, as it compares the learned memory field with its integral definition. Here, discretization is intentional and localized: it does not approximate the governing physics, but only provides accurate numerical evaluation of the memory integral needed for consistency. Thus, MINPO separates the continuous physics loss from the discrete memory-evaluation step, ensuring both accuracy and computational efficiency.

\section{Results and Discussion}\label{sec:results}

To evaluate the robustness of the MINPO, we compare its performance against A-PINN~\cite{yuan2022pinn}, fPINN~\cite{pang2019fpinns}, and their respective KAN variants as detailed in~\ref{A-PIKANN} and~\ref{fPIKANN}. Additionally, we assess it alongside conventional numerical techniques, such as Finite Difference (FD). To quantitatively assess the predictive accuracy of each model, we define the following relative error measure,
\begin{equation}
    \mathcal{E}(\hat{\nu}) =\frac{\Vert\hat{\nu} - \nu\Vert}{\Vert\nu\Vert},
\end{equation}
where \(\hat{\nu}\) denotes the neural network prediction of the corresponding quantity \(\nu\).
The notation \(\Vert\cdot\Vert\) represents the \(\mathcal{L}^2\) norm when \(\nu\) is a function such as \(u\) or \(\mathcal{M}[u]\), and the absolute value when \(\nu\) is a scalar parameter (e.g., \(\kappa)\).
The symbol ``hat" indicates an estimated or learned approximation of the true physical quantity.

\subsection{Experiment I: Nonlinear Volterra IDE}\label{ExpI:Volterra}

The first experiment is designed to assess the ability of MINPO to handle nonlinear Volterra-type memory integrals over long temporal domains, in both forward and inverse settings. To this end, we consider a representative model of wave propagation in complex materials, where nonlocal and history-dependent effects are captured by a one-dimensional nonlinear Volterra IDE~\cite{yuan2022pinn},
\begin{equation}\label{Eq:VolterraIDE}
\frac{du(t)}{dt} + u(t) = \kappa \int_0^t e^{(\tau-t)} u(\tau)\, d\tau, \quad t \in [0, A],
\end{equation}
where $u$ is the unknown field variable, $\kappa$ is a scalar parameter controlling the strength of the memory term, and $A$ denotes the length of the temporal domain.
Given the initial condition $u(0) = 1$, the analytical solution for Eq.~\eqref{Eq:VolterraIDE} is $u(t) = e^{-t} \cosh(\sqrt{\kappa}\,t)$.

Following the general formulation in  Eq.~\eqref{eq:general-IDE} with $\lambda_{\alpha}=0$, the operators in this experiment are defined as,
\begin{equation}
    \mathcal{N}[u](t) =  u(t), \qquad
\mathcal{M}[u](t) = \int_0^t e^{(\tau- t)} u(\tau)\, d\tau ,
\end{equation}
where $\mathcal{M}[u]$ captures the exponentially weighted memory kernel that models hereditary effects in the medium.
Since $e^{(\tau- t)}$ is smooth and nonvanishing for \(\tau \ge t\) while vanishing for \(\tau < t\) (i.e., it is a causal kernel), the associated Volterra operator is injective and has a unique continuous inverse~\cite{tricomi1985integral, gripenberg1990volterra, brunner2017volterra}. Thus, $\mathcal{M}$ is invertible on the space of continuous functions defined on the interval \([0,A]\).
Hence, the unknown field $u$ can be explicitly recovered from the learned integral representation $\mathcal{M}_{\boldsymbol{\theta}}$ using the inverse mapping, which represents the one-dimensional reduction of the general reconstruction ansatz in Eq.~\eqref{eq:recon-ansatz} for the Volterra IDE,
\begin{equation}\label{Eq:VolterraInverse}
u_{\boldsymbol{\Theta}}(t) = \partial_t \mathcal{M}_{\boldsymbol{\theta}}(t) + \mathcal{M}_{\boldsymbol{\theta}}(t),
\end{equation}
where the time derivative $\partial_t \mathcal{M}_{\boldsymbol{\theta}}$ in the implementation is computed efficiently via automatic differentiation.
A detailed derivation of Eq.~\eqref{Eq:VolterraInverse}, including the differentiation of the Volterra operator and the resulting operator identity, is provided in~\ref{app:inv_M-I}.
Moreover, the corresponding nonlocal consistency term (Eq.~\eqref{Eq:Mloss_1}) is formulated as,
\begin{equation}\label{Eq:EpI_Mloss}
\mathcal{L}_{\mathcal{M}}(\mathcal{D}_{\mathcal{M}}; \boldsymbol{\theta})
= \text{MSE}\left(
\mathcal{M}_{\boldsymbol{\theta}}(t) - 
 \int_0^{t}\, e^{(\tau-t)} u_{\boldsymbol{\Theta}}(\tau)\, d\tau
  \right) + \text{MSE}\left(\mathcal{M}_{\boldsymbol{\theta}}(0) - 0 \right), 
\end{equation}
which enforces the learned network output $\mathcal{M}_{\boldsymbol{\theta}}$ to be consistent with the analytical form of the operator $\mathcal{M}[u]$.
The second term in Eq.~\eqref{Eq:EpI_Mloss} enforces the natural initial condition \(\mathcal{M}(0)=0\), since the Volterra integral vanishes at \(t=0\).
Here, the integral in the nonlocal consistency loss~\eqref{Eq:EpI_Mloss} is computed using a Gauss–Legendre quadrature rule~\cite{swarztrauber2003computing, laurie2001computation}.
Because this term is not part of the governing IDE loss, such a low-cost discretization suffices to accurately compute the memory integral, keeping the additional computational cost negligible compared with conventional approaches. This reinforces MINPO’s separation between the fully continuous IDE loss and the discrete memory-evaluation step, ensuring both efficiency and accuracy.

\subsubsection{The forward problem}\label{ExpI:Volterra_Forward}

\begin{figure}[h!]
    \centering
    \includegraphics[width=1.0\linewidth]{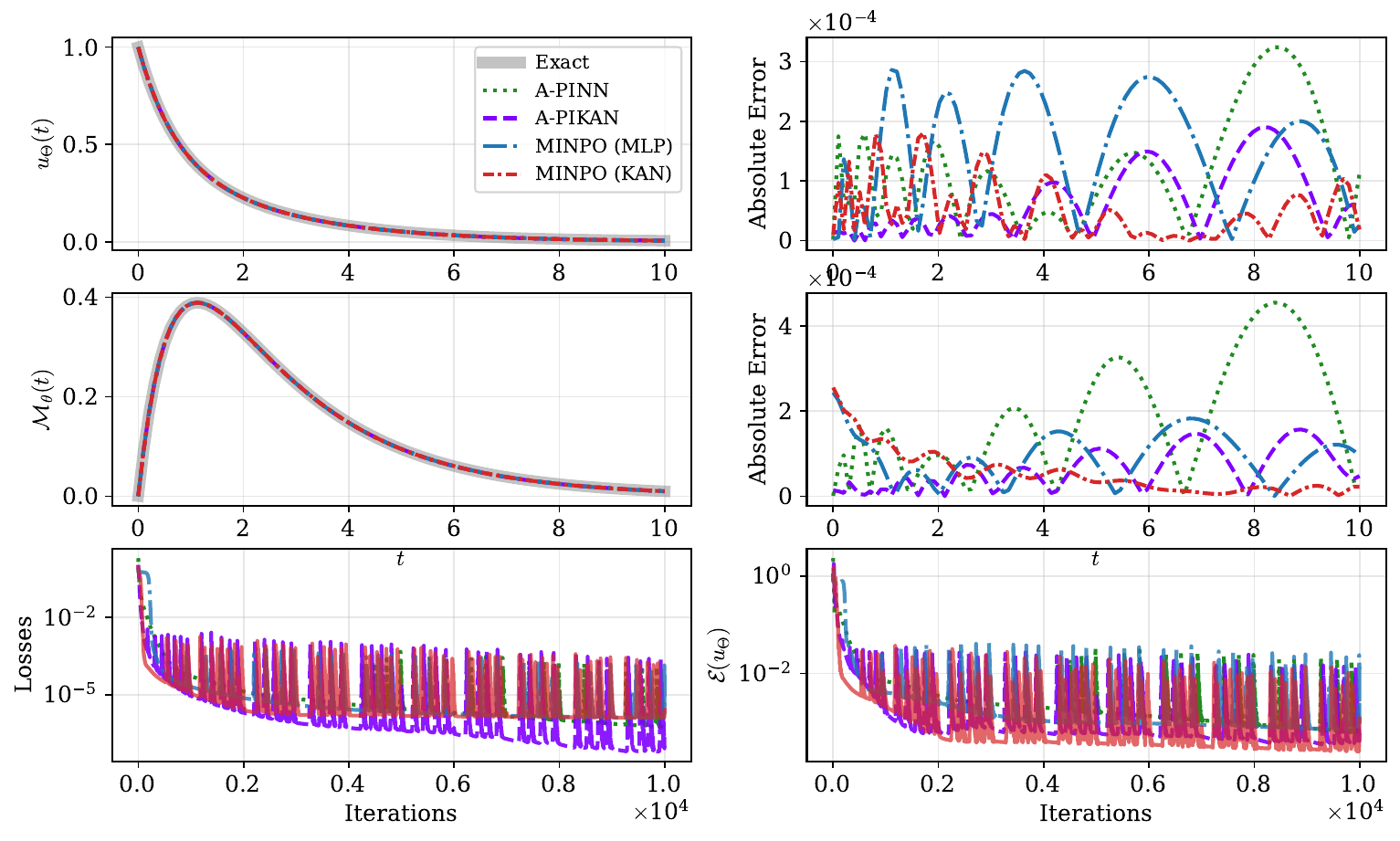}
    \caption{Forward reconstruction results for Experiment I (Exp.~\ref{ExpI:Volterra}) using MINPO (KAN/MLP) and adaptive A-PINN/A-PIKAN baselines. The KAN models use 3 layers with 15 neurons and Chebyshev degree 4, and the MLP models consist of 3 layers with 33 neurons and the \(\tanh\) activation function. All models are trained with 2400 residual points.
Across the accuracy metrics, solution reconstruction, and memory-operator recovery, MINPO-KAN achieves the best performance, while all methods exhibit comparable convergence behavior.
}
    \label{fig:expI_forward_1}
\end{figure}


In the forward setting of Experiment I (Exp.~\ref{ExpI:Volterra}), the parameter \(\kappa\) is assumed to be known, and the objective is to accurately reconstruct the solution \(u\) of the nonlinear Volterra IDE,  Eq.~\eqref{Eq:VolterraIDE}. We evaluate the ability of MINPO to learn the field \(u\) and its associated memory term directly from data, without discretizing the governing IDE. 
Figure~\ref{fig:expI_forward_1} compares MINPO (KAN/MLP) with the adaptive A-PINN/A-PIKAN baselines. A key purpose of this experiment is to assess whether the discrete treatment of memory loss in MINPO (Eq.~\eqref{Eq:EpI_Mloss}) adversely affects accuracy relative to the fully continuous loss formulation used in A-PINN and A-PIKAN.  The results indicate that this discretization does not degrade performance; instead, owing to the structure of the MINPO formulation and its non-discretized IDE loss component, the proposed method achieves accuracy that is fully competitive with, and in several metrics superior to, the adaptive baselines. 

The results in Fig.~\ref{fig:expI_forward_1} also show that MINPO, particularly the KAN-based variant, achieves the lowest maximum errors in both the reconstructed field \(u\) and the learned memory operator \(\mathcal{M}\). Specifically, MINPO (KAN) reduces the maximum error in \(u\) by about 44\% and the error in \(\mathcal{M}\) by about 66\% compared with A-PINN, and still maintains a 5\% lower error in \(u\) compared with A-PIKAN despite using a discretized memory loss, $\mathcal{L_M}$.  Interestingly, even the MLP-based MINPO, while not as accurate as the KAN version, still outperforms A-PINN, reducing its maximum error by 12\% for \(u\) and 46\% for \(\mathcal{M}\). These observations highlight that MINPO’s non-discretized IDE loss preserves accuracy, and the discretized memory term does not hinder performance.
Finally, the convergence curves shown in Fig.~\ref{fig:expI_forward_1} illustrate that all methods reach similarly low loss values, with A-PIKAN converging to the smallest residual, followed by A-PINN. In this experiment, the proposed method converges slightly more slowly but still reaches a satisfactorily low regime, confirming that its discretized memory loss does not hinder optimization and preserves the overall solution quality.

\begin{table}[h!]
\footnotesize
\centering
\caption{Results for Experiment I (Exp.~\ref{ExpI:Volterra}) in forward mode reporting the relative errors of \(u\) and the memory term for different values of \(\kappa\).
For cKAN models, a three-layer network with 15 neurons per layer and polynomial degree 4 is used, while MLP models use a three-layer network with 33 neurons per layer and the \(\tanh\) activation function.
All models are trained with 2400 residual points using 10,000 Adam iterations followed by L-BFGS optimization.
MINPO explicitly learns the memory operator, whereas A-PIKAN, and A-PINN directly approximate the IDE solution.}
\renewcommand{\arraystretch}{1.2}
\begin{tabularx}{\linewidth}{p{2.3cm} X X X X X}
\toprule
$\kappa$ & & MINPO (KAN) & MINPO (MLP) & A-PIKAN & A-PINN\\
\midrule
1.0 & $\mathcal{E}(u_{\boldsymbol{\Theta}})$ & \boldsymbol{$1.49\times 10^{-4}$} & $4.05\times 10^{-4}$& $3.27\times 10^{-4}$ & $4.28\times 10^{-4}$\\
 & $\mathcal{E}(\mathcal{M}_{\boldsymbol{\theta}})$ & \boldsymbol{$1.89\times 10^{-4}$} & $3.83\times 10^{-4}$& $2.85\times 10^{-4}$ & $2.22\times 10^{-4}$\\
 \midrule
0.8 & $\mathcal{E}(u_{\boldsymbol{\Theta}})$ & \boldsymbol{$2.07\times 10^{-4}$} & $6.39\times 10^{-4}$& $2.62\times 10^{-4}$ & $7.37\times 10^{-4}$\\
 & $\mathcal{E}(\mathcal{M}_{\boldsymbol{\theta}})$ & \boldsymbol{$2.50\times 10^{-4}$} & $5.96\times 10^{-4}$& $3.09\times 10^{-4}$ & $5.56\times 10^{-4}$\\
 \midrule
0.5& $\mathcal{E}(u_{\boldsymbol{\Theta}})$ & \boldsymbol{$1.17\times 10^{-4}$} & $5.85\times 10^{-4}$& $2.93\times 10^{-4}$ & $5.86\times 10^{-4}$\\
 & $\mathcal{E}(\mathcal{M}_{\boldsymbol{\theta}})$ & \boldsymbol{$3.43\times 10^{-4}$} & $6.24\times 10^{-4}$& $3.69\times 10^{-4}$ & $7.59\times 10^{-4}$\\
\midrule
0.3 & $\mathcal{E}(u_{\boldsymbol{\Theta}})$ & \boldsymbol{$2.63\times 10^{-4}$} & $6.87\times 10^{-4}$& $3.43\times 10^{-4}$ & $5.88\times 10^{-4}$\\
 & $\mathcal{E}(\mathcal{M}_{\boldsymbol{\theta}})$ & \boldsymbol{$3.95\times 10^{-4}$} & $5.85\times 10^{-4}$& $4.22\times 10^{-4}$ & $1.22\times 10^{-3}$\\
\bottomrule
\end{tabularx}
\label{tab:Exam1-forward}
\end{table}

Table~\ref{tab:Exam1-forward} presents a quantitative comparison between the proposed MINPO method and two baseline approaches for the forward Volterra IDE across four values of \(\kappa\). 
When averaged across all \(\kappa\), MINPO (KAN) reduces the error in \(u\) by approximately 35-45\% compared with A-PIKAN and by roughly 55-65\% relative to A-PINN.
For the memory term, the corresponding improvements are on the order of 25-40\% over A-PIKAN and 45-60\% over A-PINN.
The MLP version of MINPO, though less accurate than its KAN counterpart, still delivers systematically better performance than A-PINN for both error metrics, with average reductions of about 10-20\% for \(u\) and 35-50\% for \(\mathcal{M}\).
Across all tested values of \(\kappa\), Table~\ref{tab:Exam1-forward} shows that the proposed MINPO method, especially its KAN-based variant, consistently achieves the most accurate reconstructions of both the solution field and the memory operator, even though it employs a discretized memory loss, however not in the IDE loss. This confirms that MINPO preserves the continuous physics of the IDE while avoiding the computational cost of repeated integral evaluations in the IDE loss. Moreover, the MLP version still outperforms A-PINN in most cases, indicating that explicitly learning the memory operator provides clear benefits regardless of the encoder.

\subsubsection{The inverse problem}\label{ExpI:Volterra_Inverse}

In the second part of Experiment I (Exp.~\ref{ExpI:Volterra}), we consider the inverse problem associated with the nonlinear Volterra IDE, Eq.~\eqref{Eq:VolterraIDE}. Here, the memory strength \(\kappa\) is unknown and must be inferred simultaneously with the field \(u\). The inverse formulation is more challenging than the forward case because \(\kappa\) influences the global amplitude and growth rate of the solution, making the recovery of both \(u\) and \(\kappa\) strongly coupled. Consequently, inaccuracies in the reconstructed field can directly affect the estimation of \(\kappa\), and vice versa~\cite{aster2018parameter, yuan2022pinn}. Since this is an inverse problem, the reconstruction is performed using a set of measurement data randomly sampled from the exact solution.

\begin{table}[h!]
\footnotesize
\centering
\caption{Results for Experiment I (Exp.~\ref{ExpI:Volterra}) in inverse mode showing the relative errors of \(u\) and the memory term for different values of \(\kappa\).
For cKAN models, a three-layer network with 15 neurons per layer and polynomial degree 3 is used, while MLP models use a three-layer network with 30 neurons per layer and the \(\tanh\) activation function.
All models are trained with 2000 residual points and 10 measurement data using 10,000 Adam iterations followed by L-BFGS optimization.
MINPO accurately identifies both the unknown field and the memory parameter \(\kappa\), while  A-PIKAN and A-PINN show reduced accuracy in recovering the inverse solution.}
\renewcommand{\arraystretch}{1.2}
\begin{tabularx}{\linewidth}{p{2.3cm} X X X X X}
\toprule
$\kappa$ & & MINPO (KAN) & MINPO (MLP) & A-PIKAN & A-PINN\\
\midrule
1.0 & $\mathcal{E}(u_{\boldsymbol{\Theta}})$ & \boldsymbol{$1.21\times 10^{-4}$} & $3.06\times 10^{-4}$& $1.77\times 10^{-4}$ & $4.59\times 10^{-3}$\\
 & $\mathcal{E}(\mathcal{M}_{\boldsymbol{\theta}})$ & \boldsymbol{$1.05\times 10^{-4}$} & $2.66\times 10^{-4}$& $2.24\times 10^{-4}$ & $3.52\times 10^{-3}$\\
  & $\mathcal{E}(\kappa)$ & \boldsymbol{$3.24\times 10^{-5}$} & $5.44\times 10^{-5}$& $4.63\times 10^{-5}$ & $1.99\times 10^{-2}$\\
 \midrule
0.8 & $\mathcal{E}(u_{\boldsymbol{\Theta}})$ & \boldsymbol{$1.80\times 10^{-4}$} & $6.96\times 10^{-4}$& $1.60\times 10^{-4}$ & $8.81\times 10^{-3}$\\
 & $\mathcal{E}(\mathcal{M}_{\boldsymbol{\theta}})$ & \boldsymbol{$2.22\times 10^{-4}$} & $5.41\times 10^{-4}$& $3.41\times 10^{-4}$ & $8.78\times 10^{-3}$\\
 & $\mathcal{E}(\kappa)$ & \boldsymbol{$5.51\times 10^{-6}$} & $2.01\times 10^{-4}$& $8.60\times 10^{-5}$ & $1.03\times 10^{-1}$\\
 \midrule
0.5& $\mathcal{E}(u_{\boldsymbol{\Theta}})$ & \boldsymbol{$1.65\times 10^{-4}$} & $9.34\times 10^{-4}$& $2.98\times 10^{-4}$ & $4.86\times 10^{-3}$\\
 & $\mathcal{E}(\mathcal{M}_{\boldsymbol{\theta}})$ & \boldsymbol{$2.95\times 10^{-4}$} & $7.07\times 10^{-4}$& $3.91\times 10^{-4}$ & $6.23\times 10^{-3}$\\
 & $\mathcal{E}(\kappa)$ & \boldsymbol{$5.69\times 10^{-5}$} & $2.73\times 10^{-4}$& $1.09\times 10^{-4}$ & $1.35\times 10^{-1}$\\
\midrule
0.3 & $\mathcal{E}(u_{\boldsymbol{\Theta}})$ & \boldsymbol{$1.89\times 10^{-4}$} & $1.92\times 10^{-3}$& $2.54\times 10^{-4}$ & $2.94\times 10^{-3}$\\
 & $\mathcal{E}(\mathcal{M}_{\boldsymbol{\theta}})$ & \boldsymbol{$2.50\times 10^{-4}$} & $1.52\times 10^{-3}$& $7.93\times 10^{-4}$ & $5.73\times 10^{-3}$\\
  & $\mathcal{E}(\kappa)$ & $4.96\times 10^{-4}$ & $1.52\times 10^{-3}$& \boldsymbol{$4.22\times 10^{-4}$} & $3.91\times 10^{-1}$\\
\bottomrule
\end{tabularx}
\label{tab:Exam1-Inverse}
\end{table}

Table~\ref{tab:Exam1-Inverse} reports the performance of the four approaches on the inverse Volterra problem and demonstrates a clear advantage for the proposed MINPO framework. 
MINPO (KAN) achieves the lowest relative error in almost every regime, reducing the average relative error of the memory term by about 50-70\% compared to A-PIKAN and by over one order of magnitude compared to A-PINN. Its estimation of \(\kappa\) is particularly robust: MINPO (KAN) achieves an average improvement of approximately 60–80\% relative to A-PIKAN and recovers \(\kappa\) with errors that are three to four orders of magnitude smaller than A-PINN. Even MLP-based MINPO, despite its simpler encoder, consistently outperforms A-PINN in all metrics and remains competitive with A-PIKAN. These results highlight that learning the memory operator directly provides a stable mechanism for disentangling the coupled inverse variables, enabling the proposed method to recover both \(u\) and \(\kappa\) with high fidelity while preserving the underlying structure of the IDE.

\begin{figure}[h]
    \centering
    \includegraphics[width=0.9\linewidth]{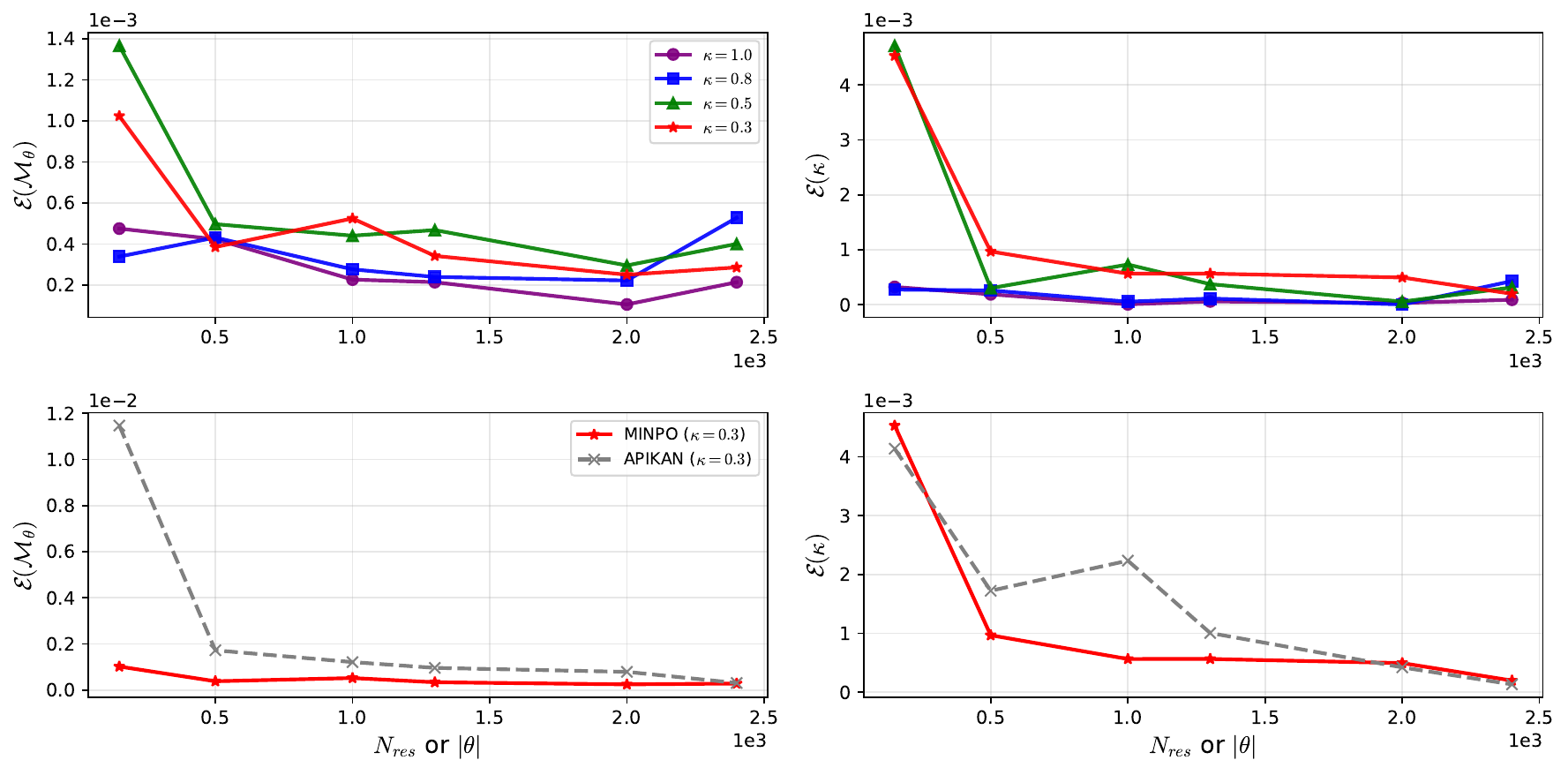}
    
    \caption{Performance of the proposed MINPO method on the inverse formulation of  Experiment I (Exp.~\ref{ExpI:Volterra}). The top row shows the performance of MINPO (KAN) across all tested values of \(\kappa\). The bottom row compares MINPO with A-PIKAN for $\kappa=0.3$. In all cases, the KAN architecture is selected so that the number of available training samples $N_{\text{res}}$ is approximately equal to the total number of trainable parameters $\vert \boldsymbol{\theta} \vert$, ensuring a balanced and fair comparison across methods.}
    \label{fig:expI_inv}
\end{figure}

Figure~\ref{fig:expI_inv} illustrates how reconstruction accuracy improves in the inverse Volterra problem as the number of residual points, and equivalently the total number of trainable parameters, increases. In the top row, which reports MINPO (KAN) efficacy across all \(\kappa\), both error metrics exhibit stable and monotonic decay. For networks trained with more than 500 residual points, the relative error of the learned memory operator falls below 0.06\%. Likewise, the relative error in \(\kappa\) reaches values below 0.01\%. These trends confirm that the proposed method reliably identifies both the memory structure and the governing parameter from limited measurements.
The bottom row compares MINPO with A-PIKAN for a representative case. Although both methods eventually converge when large networks are used, the difference is pronounced in the low-data regime. For fewer than 500 residual points, the relative error in the memory operator is approximately 0.1\% for MINPO but around 1.2\% for A-PIKAN. A-PIKAN also shows higher sensitivity to model size, particularly in recovering \(\kappa\).
Figure~\ref{fig:expI_inv} highlights that MINPO achieves accurate and robust inverse reconstructions across different network capacities, effectively capturing both the nonlinear Volterra memory operator and the unknown scaling parameter. These results underscore the capability of the proposed operator-informed framework to model nonlocal dynamics even under limited data or network resources.

\subsection{Experiment II: Three-Dimensional Nonlocal IDE}\label{ExpII:3DIDE}

This experiment is designed to evaluate the performance of the proposed MINPO framework in higher-dimensional settings, where both the dimensionality of the domain and the number of nested integrals are increased.
We consider a three-dimensional integro-differential equation defined as,  
\begin{equation}\label{Eq:3DIDE}
\left(\frac{\partial}{\partial {t}} + \frac{\partial}{\partial {x_1}} + \frac{\partial}{\partial {x_2}} \right)\, u(\boldsymbol{\xi})  = u(\boldsymbol{\xi}) + \int_{0}^{x_2}\int_{0}^{x_1}\int_{0}^{t}
e^{(\tau-t)}\, u(y_1, y_2, \tau)\, d\tau\, dy_1\, dy_2 + f(\boldsymbol{\xi}),
\end{equation}
where $\boldsymbol{\xi}=(x_1,x_2,t)\in \Omega_{\boldsymbol{\xi}} = [0,1]^3$, 
$u$ is the scalar field, and the integral term accounts for the accumulated nonlocal contributions over the temporal dimension and both spatial directions.
The source term $f$ is determined analytically to ensure that the exact solution, 
\begin{equation}
u(\boldsymbol{\xi}) = t \sin(x_1)\cos(x_2), \qquad \boldsymbol{\xi} = (x_1, x_2, t) \in \Omega_{\boldsymbol{\xi}}^{-},
\end{equation}
satisfies Eq.~\eqref{Eq:3DIDE}.
Following the representation in Eq.~\eqref{eq:general-IDE}, the operators are defined as,
\begin{equation}
\mathcal{N}[u](\boldsymbol{\xi}) =
\left( 1 - \frac{\partial}{\partial {x_1}} - \frac{\partial}{\partial {x_2}}  \right)\, u (\boldsymbol{\xi}),
\qquad
\mathcal{M}[u](\boldsymbol{\xi})
 =
\int_{0}^{x_2}\int_{0}^{x_1}\int_{0}^{t}
e^{(\tau-t)}\, u(y_1, y_2, \tau)\, d\tau\, dy_1\, dy_2,
\end{equation}
where \(\mathcal{N}[u]\) governs the local transport dynamics, and \(\mathcal{M}[u]\) encodes the three-dimensional cumulative memory effects of the field through a smooth and causal kernel. 
An important property of the operator $\mathcal{M}$ is that it vanishes whenever any of the coordinates $x_1$, $x_2$, or the time variable $t$ equals zero, i.e.,
\begin{equation}
    \mathcal{M} = 0 \quad \text{if} \quad x_1 = 0 \ \text{or}\  x_2 = 0 \ \text{or}\  t = 0.
\end{equation}

To ensure that the learned representation \(\mathcal{M}_{\boldsymbol{\theta}}\) satisfies this boundary property exactly, we impose it through a hard constraint in the network design. Specifically, the network output is multiplied by the product \(x_1\, x_2\, t\), such that,
\begin{equation}
    \mathcal{M}_{\boldsymbol{\theta}}(\boldsymbol{\xi}) = x_1\, x_2\, t \, \tilde{\mathcal{M}}_{\boldsymbol{\theta}}(\boldsymbol{\xi}),
\end{equation}
where \(\tilde{\mathcal{M}}_{\boldsymbol{\theta}}\) denotes the unconstrained output of the neural network.
This formulation guarantees that \(\mathcal{M}_{\boldsymbol{\theta}}\) naturally satisfies the required zero-boundary condition at the coordinate planes, while preserving smoothness and learning flexibility in the interior of the domain.
Since the integral kernel \(e^{(\tau-t)}\) is regular and monotonic, the operator \(\mathcal{M}\) is invertible, and the recovery of \(u\) follows from the higher-dimensional specialization of the reconstruction ansatz in Eq.~\eqref{eq:recon-ansatz}, leading to,
\begin{equation}\label{Eq:3DInverse}
u_{\boldsymbol{\Theta}}(\boldsymbol{\xi}) = 
\frac{\partial^3 \mathcal{M}_{\boldsymbol{\theta}}}{\partial t\, \partial x_1\, \partial x_2}
+
\frac{\partial^2 \mathcal{M}_{\boldsymbol{\theta}}}{\partial x_1\, \partial x_2},
\qquad \boldsymbol{\xi} = (x_1, x_2, t) \in \Omega_{\boldsymbol{\xi}}^{-},
\end{equation}
for which all partial derivatives in are efficiently computed using automatic differentiation, ensuring consistency between the learned integral representation and the governing IDE. A detailed derivation of this inverse relation is provided in~\ref{app:inv_M-3D}. The nonlocal consistency term used in the total loss function (Eq.~\eqref{Eq:Mloss_1}) is formulated as,
\begin{equation}\label{Eq:ExII_MemoryLoss}
\mathcal{L}_{\mathcal{M}}(\mathcal{D}_{\mathcal{M}}; \boldsymbol{\theta})
=
\text{MSE}\left(
\mathcal{M}_{\boldsymbol{\theta}} -
\int_{0}^{x_2}\int_{0}^{x_1}\int_{0}^{t}
e^{(\tau-t)}\, u_{\boldsymbol{\Theta}}(y_1, y_2, \tau)\, d\tau\, dy_1\, dy_2
\right).
\end{equation}

In Eq.~\eqref{Eq:ExII_MemoryLoss}, the triple integral appearing in the nonlocal consistency loss is evaluated numerically using an \(N_I\)-point Gauss-Legendre quadrature rule~\cite{swarztrauber2003computing, laurie2001computation}. This quadrature is used only within the loss term to enforce agreement between the learned representation \(\mathcal{M}_{\boldsymbol{\theta}}\) and the analytical form of the memory operator. Crucially, as it can be seen  MINPO does not discretize the governing IDE itself; the model retains a fully continuous formulation, and the integral kernel remains implicitly encoded through the learned operator. 

\begin{table}[h!]
\footnotesize
\centering
\caption{Structural and accuracy comparison between MINPO (KAN) and A-PIKAN for the three-dimensional nonlocal IDE in Experiment II (Exp.~\ref{ExpII:3DIDE}).
The models employ a 3-layer KAN architecture with 10 neurons per layer and Chebyshev degree (k=3), using 1000 residual points during training.}
\label{tab:Exam2}
\renewcommand{\arraystretch}{1.25}

\begin{tabularx}{\linewidth}{p{2.5cm} p{3cm} X X}
\toprule
Category & Item & MINPO (KAN) & A-PIKAN \\
\midrule

\multirow{4}{*}{\textbf{Structure}}
 & Unknowns & $\mathcal{M}, u$ & $u, v_1, v_2, v_3$ \\
 & IDE residual & Single IDE & Four coupled ODE-PDEs\\
 & Memory handling & Explicit (via $\mathcal{L}_{\mathcal{M}}$) & Implicit (via $v_i$) \\
 & Loss terms & Residual + data + memory & Residual + data \\

\midrule

\multirow{2}{*}{\textbf{Accuracy}}
 & $\mathcal{E}(u_{\boldsymbol{\Theta}})$ & $4.49 \times 10^{-3}$ & $8.98 \times 10^{-3}$ \\
 & $\mathcal{E}(\mathcal{M}_{\boldsymbol{\theta}})$ & $1.46 \times 10^{-3}$ & $9.87 \times 10^{-2}$ \\

\bottomrule
\end{tabularx}
\end{table}

The three-dimensional nonlocal IDE in Experiment II (Exp.~\ref{ExpII:3DIDE}) is solved using MINPO (KAN), and the resulting structural and accuracy metrics are reported in Table~\ref{tab:Exam2}. For comparison, results from A-PIKAN, whose full formulation is provided in~\ref{A-PIKANN}, are also included, focusing on the core differences relevant to evaluating operator learning. MINPO learns the solution field \(u\) and the memory operator \(\mathcal{M}\), enforcing the IDE directly without auxiliary states or problem-specific reformulations. A-PIKAN, by contrast, embeds memory implicitly through three auxiliary fields \((v_1, v_2, v_3)\), which expand the model dimension and introduce a coupled ODE–PDE system. This structural overhead becomes especially critical for high-dimensional and nested memory integrals, where compact operator representations directly influence robustness and computational efficiency.
Regarding accuracy, both approaches recover the solution field with comparably high precision; MINPO achieves a relative error about 50\% lower than A-PIKAN. However, MINPO exhibits a decisive advantage in learning the memory operator, reducing the relative error by roughly 98\%, nearly an order of magnitude. This improvement reflects the effectiveness of MINPO’s explicit consistency loss, which discretizes the integral only as an external regularization step while preserving the continuous physics of the IDE. 
Taken together, Table~\ref{tab:Exam2} shows that MINPO handles the dimensional complexity of the three-dimensional operator more efficiently and provides substantially more accurate memory reconstruction, consistent with the overarching goals of the framework.

\begin{figure}[h!]
    \centering
    \includegraphics[width=1.0\linewidth]{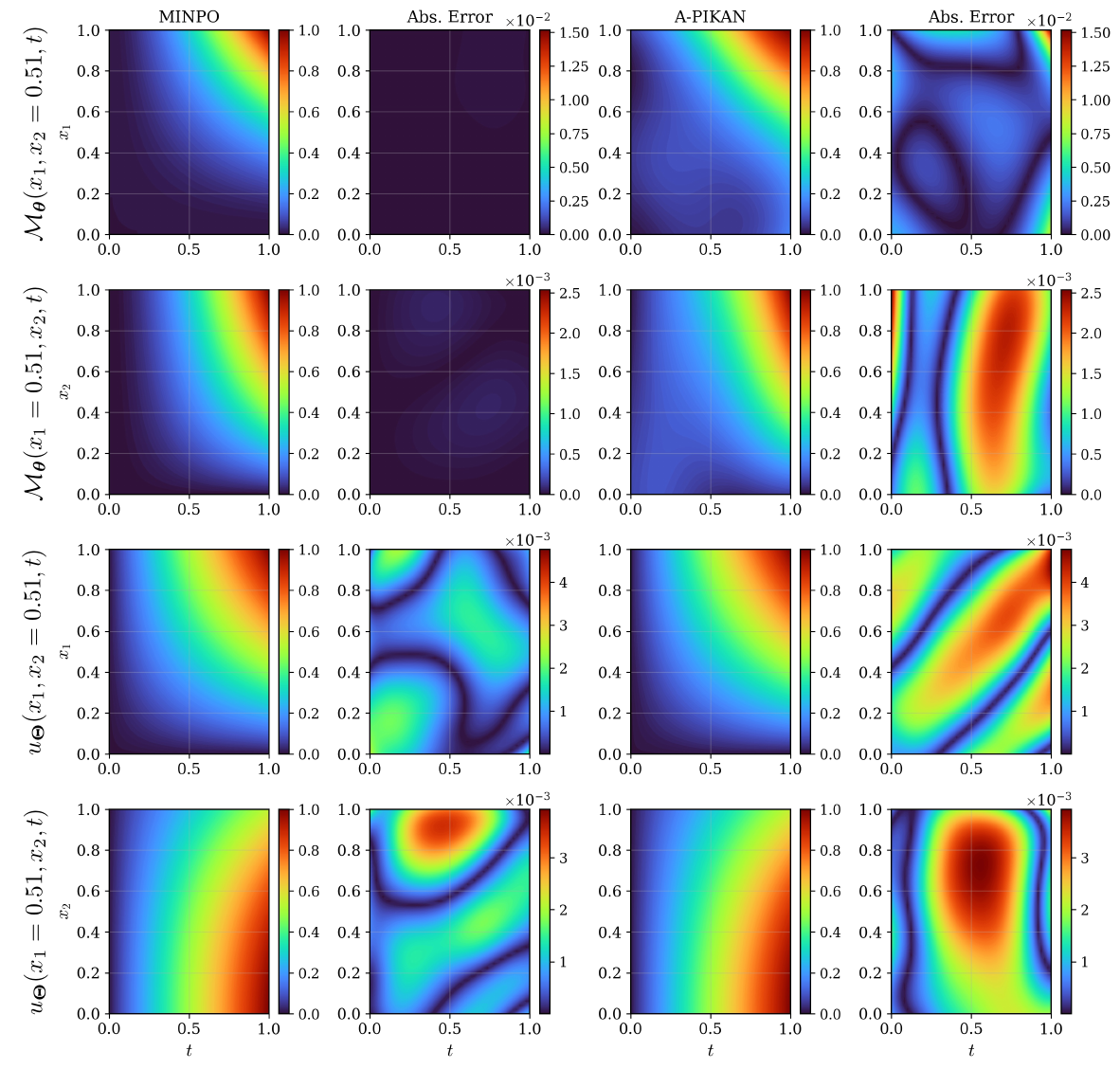}
    \caption{Comparison of the predicted memory operator \(\mathcal{M}[u]\) and solution field \(u\), together with their absolute errors, for Experiment II (Exp.~\ref{ExpII:3DIDE}). Both MINPO (KAN) and A-PIKAN employ a 3-layer KAN architecture with 10 neurons per layer and Chebyshev degree \(k=3\), trained using 1000 residual points; for MINPO, the nonlocal consistency loss uses \(N_I = 10\) quadrature samples. Across all displayed slices, MINPO yields substantially smaller errors in the learned memory operator while preserving solution accuracy comparable to that of A-PIKAN, highlighting its capability in recovering high-dimensional nonlocal operators.
}
    \label{fig:ExII_minpo-kan}
\end{figure}

We additionally compare MINPO (KAN) and A-PIKAN across two representative slices of the three-dimensional domain, as shown in Fig.~\ref{fig:ExII_minpo-kan}, which enables a detailed examination of their behavior along different spatial sections.
In the first two rows, which display slices of the memory operator \(\mathcal{M}[u]\), MINPO produces reconstructions that remain smooth and consistent across both \((x_1,0.5,t)\) and \((0.5,x_2,t)\) planes, while the corresponding A-PIKAN panels show visible distortions and larger spatiotemporal variations in the error. 
This demonstrates that MINPO captures the high-dimensional memory structure with 
substantially greater fidelity.
The third and fourth rows correspond to slices of the solution field \(u\).
In these slices, both models successfully recover the target solution and produce visually similar spatial--temporal patterns, indicating stable solution learning in both cases. However, MINPO consistently attains lower maximum errors across the examined slices. 
These results indicate that, beyond visual similarity, MINPO provides a more accurate and robust approximation of the solution field \(u\), particularly in regions where peak errors dominate the overall accuracy assessment.
These slice-wise comparisons make clear that MINPO not only delivers a decisive advantage in reconstructing the memory operator, but also maintains consistently superior performance in estimating the solution field \(u\), even when both methods appear visually comparable.

\begin{figure}[h!]
    \centering
    \includegraphics[width=1.0\linewidth]{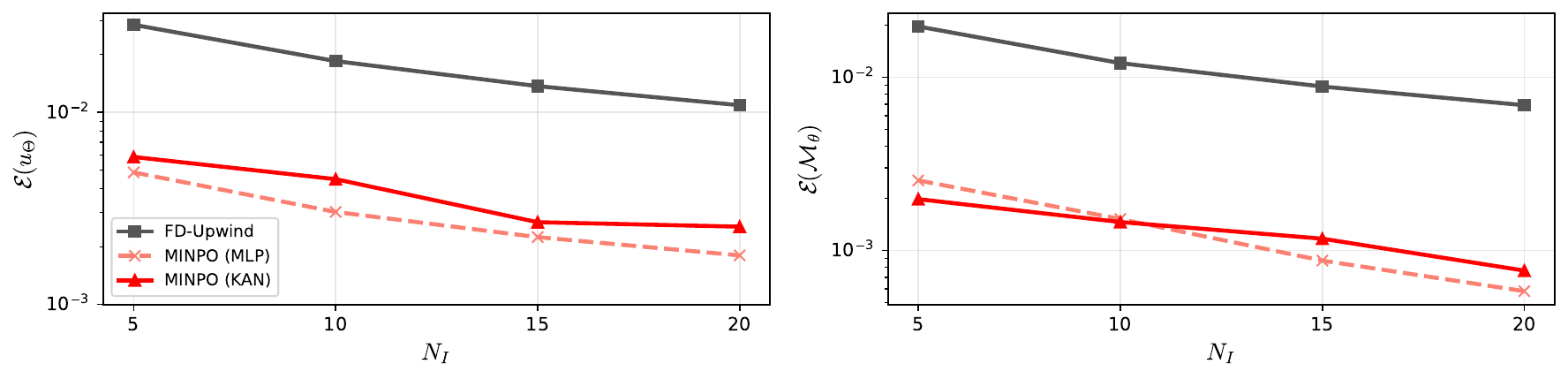}
    \caption{Relative errors versus the \(N_I\)-point Gauss–Legendre quadrature rule for Experiment II (Exp.~\ref{ExpII:3DIDE}).
MINPO (KAN) uses a 3-layer, 10-neuron architecture with Chebyshev degree 3, and MINPO (MLP) uses a 3-layer, 21-neuron network; all are trained with \(N_{\text{res}} = 1000\). Finite-difference baselines (forward and upwind) use spatial grids \(N_x^3 = 10^3, 15^3, 20^3,\) and \(25^3\). MINPO maintains higher accuracy at lower cost even with modest quadrature resolution, outperforming FD across all tests.}
    \label{fig:ExII_minpo-FD}
\end{figure}

To provide a fair comparison, we also implement a fully discrete finite-difference solver~\cite{zhao2006compact} that approximates all differential operators on a uniform three-dimensional grid and evaluates the nonlocal memory term via Gauss–Legendre quadrature. The resulting linear system is solved using a Picard–Jacobi iteration~\cite{junkins2013picard, tafakkori2020jacobi,} with initial and Dirichlet boundary conditions.
The performance comparison underscores a decisive advantage for the MINPO operator-learning approach. 
As illustrated in Fig.~\ref{fig:ExII_minpo-FD}, both MINPO variants achieve significantly lower relative errors for both the solution field \(u\) and the memory operator \(\mathcal{M}[u]\) across all quadrature resolutions \(N_I\). Specifically, for the solution field at the finest quadrature (\(N_I=20\)), MINPO (KAN) reduces the solution error by 76.7\%, while MINPO (MLP) achieves an even larger reduction of 83.5\% relative to the FD baseline. 
The improvement is even more pronounced for the memory operator: MINPO (KAN) decreases the error by 88.9\%, and MINPO (MLP) by 91.6\%, compared to the FD solver at the same quadrature resolution.
Notably, this superior accuracy is achieved using only \(N_{\text{res}} = 1000\) collocation points for training. In contrast, the FD solver requires a substantially finer spatial discretization to alleviate its inherent truncation error. Even after refining the grid to \(25^3 = 15{,}625\) nodes, approximately 15.6 times denser than the number of MINPO training points, the FD error remains significantly higher than that obtained by MINPO with modest quadrature.
Importantly, these accuracy gains are accompanied by substantial computational savings. Based on the reported runtimes, measured using MINPO with \(N_I = 5\) and the FD upwind solver with \(N_I = 20\), MINPO (MLP) reduces the computational time by a factor of approximately 15.3, while MINPO (KAN) achieves a speedup of about 4.7 relative to the FD baseline, despite delivering superior accuracy.
These results demonstrate that MINPO's strategy of preserving continuous physics and discretizing only a single loss term enables highly accurate operator recovery with minimal data. The FD solver, constrained by grid-dependent discretization and numerical diffusion, necessitates a vastly denser mesh to partially close the accuracy gap, highlighting the data efficiency and robustness of the neural operator paradigm for high-dimensional problems with nested integrals.

\subsection{Experiment III: Time-Fractional Transport Dynamics}
This experiment examines a case in which the temporal evolution in the unified IDE formulation Eq.~\eqref{eq:general-IDE} is governed by the fractional operator \(\mathcal{T}_\alpha\) defined in Eq.~\eqref{eq:T-operator}. The goal is to demonstrate that MINPO can accurately learn and invert intrinsic fractional temporal memory.

\subsubsection{1D time-fractional diffusion problem}\label{Sec:1Dfrac}

We consider a one-dimensional time-fractional diffusion equation~\cite{mostajeran2025solving},
\begin{equation}\label{Eq:ExamIII}
    {}^{C}\!D_t^{\alpha} u(x,t) = \Delta u(x,t) + S(x,t), \qquad (x,t)\in (0,1)\times (0,1),
\end{equation}
with homogeneous boundary and initial conditions,
\begin{equation}
    u(0,t)=u(1,t)=0,\qquad u(x,0)=0,
\end{equation} 
where \({}^{C}\!D_t^{\alpha}\) denotes the Caputo fractional derivative of order \(0<\alpha<1\), as defined in Eq.~\eqref{Eq:CaputoDeriv_1}, and \(\Delta\) is the Laplacian in the spatial variable. 
The exact analytical solution for this example expressed as $u(x,t) = t^3 \sin(\pi x)$.
Note that in this example there is no additional nonlocal memory operator \(\mathcal{M}[u]\) (i.e., \(\mathcal{M}\equiv 0)\); the temporal nonlocality is entirely encoded by the Caputo derivative.
Following the MINPO framework, we approximate the fractional derivative and its inverse by neural fields as in Eq.~\eqref{Eq:frac_def_network}. 
By Lemma~\ref{lem:CaputoIdentity} (in~\ref{fPIKANN}) the composition of these learned operators recovers the solution, and the MINPO reconstruction reduces to the specialised form of Eq.~\eqref{eq:recon-ansatz}
\begin{equation}
    u_{\boldsymbol{\Theta}}(x,t)
=
u(x,0)
+
\mathcal{J}_{\boldsymbol{\phi}}(x,t),
\end{equation}
which enables pointwise recovery of \(u\) without explicitly evaluating the fractional integral or storing the temporal history.

To ensure that the learned fractional memory field behaves as a genuine Caputo derivative, MINPO imposes a nonlocal consistency constraint. The corresponding loss term Eq.~\eqref{Eq:Mloss_1} is defined as,
\begin{equation}
    \mathcal{L}_{\mathcal{M}}(\mathcal{D}_{\mathcal{M}};\boldsymbol{\Theta})
=
\mathrm{MSE}\!\left(
\mathcal{M}_{\boldsymbol{\theta}}(x,t)
-
\frac{1}{\Gamma(1-\alpha)}
\int_{0}^{t}
\frac{\partial_{\tau} u_{\boldsymbol{\Theta}}(x,\tau)}{(t-\tau)^{\alpha}}
\,\mathrm{d}\tau
\right),
\end{equation}
where the trainable parameters are collected in \(\boldsymbol{\Theta}=\{\boldsymbol{\theta},\boldsymbol{\phi}\}\).
The temporal integral is evaluated numerically using the L1 discretization scheme~\cite{gao2014new}, which yields the standard finite-difference approximation of the Caputo derivative.
Importantly, MINPO differs from fPINN-type solvers in how numerical integration is used. In fPINN methods, the fractional differential equation is discretized directly, and numerical quadrature is built into the physics residual. In contrast, MINPO performs numerical integration only inside the consistency loss \(\mathcal{L}_{\mathcal{M}}\), while the IDE residual \(\mathcal{L}_{\text{IDE}}\) remains fully continuous. This separation keeps the governing physics free from discretization, allowing the IDE loss to be evaluated at any point in the domain and enabling the network to capture memory effects without depending on a fixed temporal grid. As a result, MINPO can attain higher accuracy, avoid resolution loss caused by coarse operator discretization, and remain flexible across a wide range of kernel dimensions and memory behaviors.

\begin{figure}
    \centering
    \includegraphics[width=1.0\linewidth]{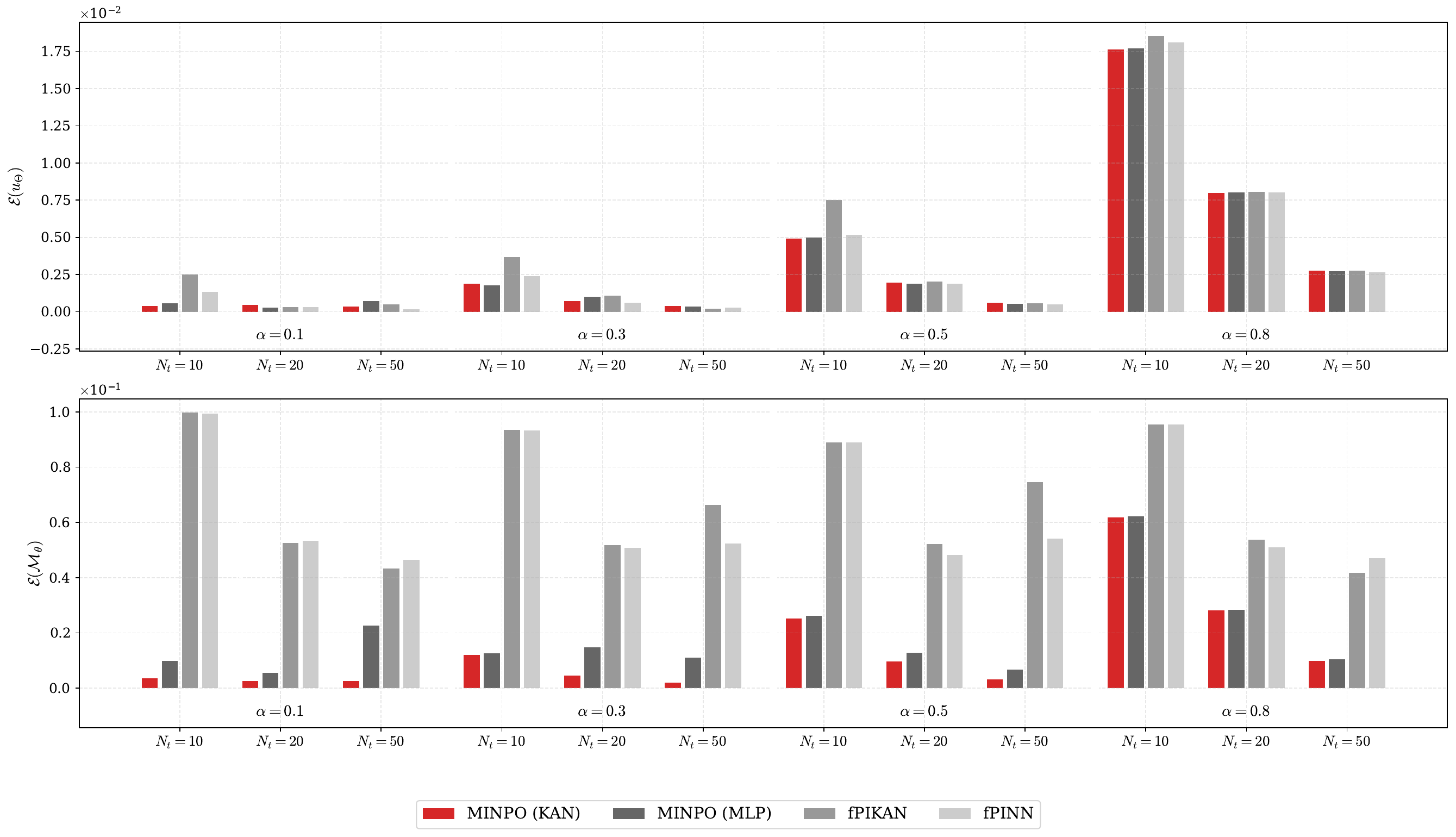}
    \caption{Relative  errors for the 1D time-fractional diffusion problem Experiment III (Exp.~\ref{Sec:1Dfrac}). Each block corresponds to a different fractional order \(\alpha\), and within each block, the three groups represent different temporal discretizations \(N_t\). KAN networks use 3 layers with 15 neurons and Chebyshev polynomials of degree 4; MLPs use 3 layers with 33 neurons and \(\tanh\) activation. Spatial sampling uses 200 points, and MINPO is trained with \(N_{\mathrm{res}}=2000\) residual constraints. }
    \label{fig:1Dfrac_1}
\end{figure}

We solve this 1D time-fractional diffusion problem using the proposed MINPO framework and, since MINPO employs the same L1 temporal discretization used in standard fractional PINN methods, we directly compare its performance against fPINN and fPIKAN, whose formulation is detailed in~\ref{fPIKANN}. The resulting relative errors for both the reconstructed solution field \(u\) and the learned memory operator approximating the Caputo derivative are reported in Fig.~\ref{fig:1Dfrac_1}. Several consistent trends emerge across these results, which reveal important differences in accuracy, robustness, and operator learning among the competing methods. A first viewpoint concerns the role of the fractional order. As \(\alpha\) increases and approaches the classical limit \(\alpha\!\to\!1\), the problem becomes intrinsically more challenging, the temporal dynamics sharpen and all models experience a rise in error (at a fixed network architecture). Even under these harder conditions, MINPO maintains the lowest reconstruction error. For example, at \(\alpha=0.8\) with \(N_t=10\), the relative error in \(u\) is around 1.7\% for MINPO (KAN), 1.7–1.8\% for MINPO (MLP), and 1.8–1.9\% for fPIKAN/fPINN. The same trend appears in the operator error, where MINPO remains around 6\%, while fPINN and fPIKAN range between 9–9.8\%. 

Another viewpoint arises by examining the influence of temporal resolution in Fig.~\ref{fig:1Dfrac_1}. Increasing \(N_t\) consistently reduces the error for all models, as finer sampling improves the supervision of the fractional dynamics. Nevertheless, MINPO already performs strongly at coarse resolutions. 
At \(\alpha=0.5\) and \(N_t=10\), MINPO reconstructs \(u\) with 35\% less error than fPIKAN and about 4\% less error than fPINN. The contrast becomes far stronger when evaluating the Caputo derivative, where MINPO achieves an error that is 72\% smaller than fPIKAN and fPINN.
This demonstrates that MINPO is far less sensitive to temporal discretization than its competitors. Another interesting viewpoint is the quality of the learned memory operator. Across nearly all (\(\alpha, N_t\)), MINPO yields substantially lower operator errors than fPINN and fPIKAN. This gap widens in the more challenging regimes (e.g., large \(\alpha\), small \(N_t\)), confirming that the pseudo-operator formulation enables MINPO to capture the underlying nonlocal structure far more effectively than residual-based approaches.
Moreover, in Fig.~\ref{fig:1Dfrac_1}, comparing KAN and MLP encoders reveals a consistent, moderate advantage for the KAN variants. MINPO (KAN) generally achieves lower error than MINPO (MLP), with the improvement being particularly visible in \(\mathcal{M}_{\boldsymbol{\theta}}\). This reflects the benefit of KAN’s polynomial-based representation for modeling nonlinear operator behavior.

These observations show that MINPO provides robust, high-accuracy reconstructions across a spectrum of fractional orders and time resolutions. This strategy, simultaneously approximating the memory operator and its reconstruction field, enables significantly better operator recovery and stronger solution accuracy than existing fractional PINN models. 
The results in Fig.~\ref{fig:1Dfrac_1} further highlight the broader capability of MINPO to reliably handle nonlocal dynamics governed by integro-differential operators, encompassing both fractional and classical PDE settings.

\begin{figure}[t]
    \centering
    \includegraphics[width=1.0\linewidth]{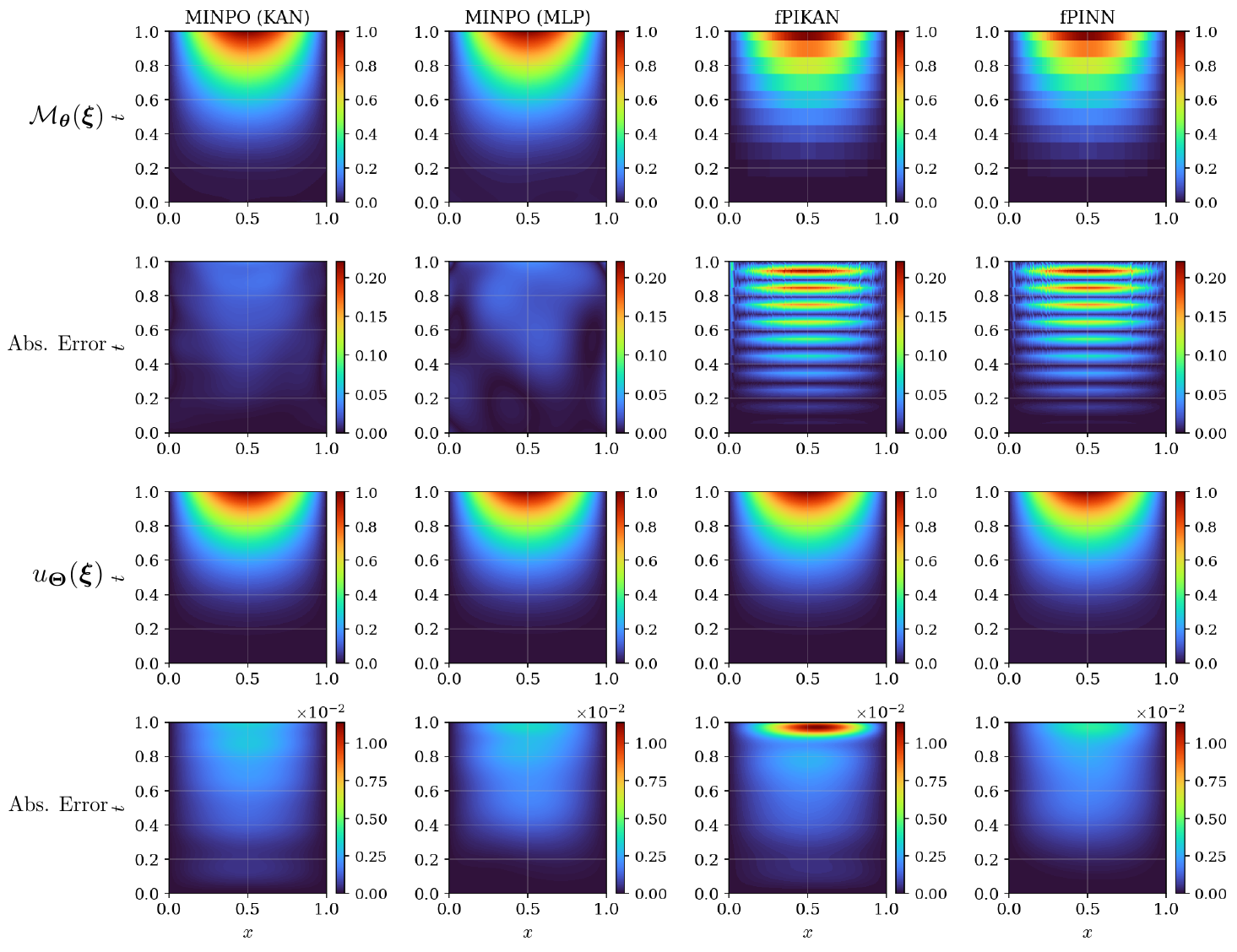}
    \caption{Comparison of the predicted Caputo memory term and solution field,  together with their absolute errors, for Experiment III (Exp.~\ref{Sec:1Dfrac}) at a fractional order of \(\alpha = 0.5\). KAN networks use 3 layers with 15 neurons and Chebyshev polynomials of degree 4; MLPs use 3 layers with 33 neurons and \(\tanh\) activation. The simulation uses \(N_t = 10\) temporal points, 200 spatial samples, and MINPO is trained with \(N_{\mathrm{res}} = 2000\) residual constraints.}
    \label{fig:1Dfrac_2}
\end{figure}

To further quantify the efficacy of MINPO for Experiment III (Exp.~\ref{Sec:1Dfrac}), we compare the results of all four methods for a representative fractional order \(\alpha = 0.5\) in Fig.~\ref{fig:1Dfrac_2}.  The most notable differences appear in the reconstruction of the Caputo memory term.
The MINPO-based models achieve smooth, fully continuous approximations of the Caputo derivative across the entire space–time domain. This behavior follows from the pseudo-operator formulation where the memory term is learned as a continuous neural operator rather than being assembled from discrete time increments. Temporal discretization enters only through the consistency loss, and therefore does not impose any piecewise structure on the learned operator. As a result, both MINPO (KAN and MLP) produce clean reconstructions with small, uniformly distributed errors.
In contrast, the memory fields reconstructed by fPINN and fPIKAN exhibit clear piecewise patterns and oscillatory artifacts. These arise directly from the L1 discrete convolution used to approximate the Caputo derivative, whose evaluation depends explicitly on the temporal mesh. Although the solution \(u\) is represented continuously in these models, the discrete structure of the governing IDE propagates into the learned memory, producing spatial–temporal irregularities and larger absolute errors.
A similar pattern is observed in the reconstruction of the solution \(u\) in Fig.~\ref{fig:1Dfrac_2}. The MINPO variants recover \(u\) with noticeably higher accuracy. The error maps show smooth, low-magnitude residuals across the domain. 
Between the two MINPO architectures, KAN-based MINPO offers a modest improvement over MLP-based MINPO, reflecting the advantage of KAN’s polynomial-based encoding in representing nonlinear operator structure.
The fPINN and fPIKAN reconstructions, on the other hand, are slightly diffused and contain sharper localized errors, again consistent with inaccuracies inherited from the discrete fractional operator.

\subsection{Discussion}

Integro–differential equations provide a unified mathematical framework for describing systems whose dynamics are influenced by memory effects or nonlocal spatial interactions. Although they arise in many applications, these equations are notoriously challenging to learn or compute, since their kernels can be singular, high-dimensional, or even unknown, and imposing standard numerical quadrature often leads to excessive computational cost and numerical instability. These persistent challenges emphasize the need for a unified framework capable of resolving both the memory operator and the solution field in continuous form, without assuming any particular structure for the underlying kernel. The MINPO framework introduced in this work addresses these issues by embedding a continuous neural representation of the nonlocal operator directly within the governing physics. This formulation eliminates explicit quadrature inside the main IDE residual, avoiding the dimensional growth typically associated with nested or multidimensional integrals and maintaining stability over long temporal memory or long range spatial horizons expressed by different non/linear kernels or nonlocal operators. 

It is demonstrated that MINPO can effectively solve forward and inverse Volterra problems, integro-differential equations with unknown coefficients, nested multi-dimensional nonlocal operators, and fractional-time models involving Caputo derivatives. 
Table~\ref{Tab:IDE_challenges}  compares this pseudo-neural approach  with other existing neural and classical approaches in addressing the fundamental challenges of integro–differential equations. 
MINPO uniquely combines the ability to handle diverse and unknown kernel structures with scalability to high-dimensional kernels, while maintaining computational efficiency and requiring minimal problem-specific manipulation. In contrast, Auxiliary-field approaches such as A-PINN and A-PIKAN reduce reliance on numerical integration and can handle certain kernel dimensions, which limits kernel diversity and necessitates substantial manual reformulation. Fractional PINN variants (fPINN/fPIKAN) accommodate power-law and fractional kernels but rely on explicit discrete approximations of the nonlocal operator, leading to increased computational cost as kernel dimensionality or memory length grows. Classical numerical methods, while flexible in kernel form, face severe scalability and efficiency limitations as kernel dimensionality increases. These comparisons highlight MINPO as a unified and scalable framework for learning and solving IDEs with complex, high-dimensional nonlocal operators.

\begin{table}[h!]
\centering
\caption{Comparison of MINPO, A-PINN/A-PIKAN, and fPINN/fPIKAN with respect to major challenges in solving IDEs.}
\label{Tab:IDE_challenges}
\footnotesize
\begin{tabular}{p{2.8cm} p{2.8cm} p{2.6cm} p{2.6cm} p{2.8cm}}
\toprule
\textbf{Model} &
\textbf{Kernel Diversity} &
\textbf{Dimensionality of the Kernal} &
\textbf{Computational Efficiency} &
\textbf{Manipulation Needed}  \\
\midrule
\addlinespace
MINPO &
$\checkmark$ &
$\checkmark$ &
$\checkmark$ &
Minimal  \\
\addlinespace
A-PINN / A-PIKAN &
$\times$ &
$\checkmark$ &
$\checkmark$ &
High  \\
\addlinespace
fPINN / fPIKAN &
$\checkmark$ &
$\checkmark$ &
$\times$ &
Minimal  \\
\addlinespace
Classical Methods &
$\checkmark$ &
$\times$ &
$\times$ &
High \\
\bottomrule
\end{tabular}
\end{table}



Although MINPO exhibits strong generalization across a wide range of nonlocal systems, its application to certain kernels may present challenges, which have not been examined in this work. specifically, when the true kernel exhibits abrupt discontinuities or extremely sharp localization~\cite{yi2015h, lischke2018fractional}, smooth neural parameterizations may require large capacity or specialized activation/basis functions to represent such features accurately. Solutions containing shocks or jump discontinuities may also challenge the continuous neural ansatz and introduce over-smoothing, a well-known limitation shared by physics-informed frameworks more broadly~\cite{pang2019fpinns, guo2022monte}. In purely data-driven inverse settings with limited or highly noisy observations, the nonlocal consistency constraint may be insufficient to uniquely identify the operator, and additional structural priors or regularization may be necessary~\cite{shan2025red}; in such cases, MINPO may approximate rather than recover the true operator.

These considerations point us toward several directions for future development. Extending MINPO to multi-term and distributed-order fractional PDEs would enable identification of systems whose memory spans a spectrum of exponents, a behavior common in complex materials~\cite{ding2021applications, failla2020advanced}. Similarly, generalizing the pseudo-operator formulation to spatial nonlocality, such as fractional Laplacians~\cite{lischke2018fractional}, tempered Lévy processes~\cite{boniece2020fractional}, or peridynamic-type interactions, would expand applicability to nonlocal diffusion, biomechanics, and electrophysiology~\cite{jian2023fractal}, where spatial nonlocal dynamics plays a dominant role. Another promising direction involves state-dependent or time-varying kernels, which arise in nonlinear rheology, electrochemical hysteresis, and biological adaptation~\cite{yu2016fractional, mainardi2022fractional, wang2022fractional, ma2023bi, zinihi2025identifying}; learning such operators would require coupling the kernel structure dynamically to the evolving solution field. Finally, integrating MINPO with symbolic or sparse-regression operator libraries \cite{ADAMSINDy} could yield interpretable memory models, enabling hybrid symbolic–neural discovery of nonlocal laws directly from data.

\section{Conclusion}\label{Sec.Conclusion}

This study introduced MINPO, a memory-informed neural pseudo-operator framework, designed to model nonlocal spatiotemporal dynamics governed by diverse integro-differential equations. The method proposes a direct neural representation of the memory operator together with a reconstruction of the solution field. MINPO allows the IDE to be enforced in its continuous form, while localizes the numerical quadrature to a minimal part of the training process through a lightweight nonlocal consistency term, only as a constraint.  The significance of MINPO thus lies in addressing long-standing structural and computational challenges in solving general IDEs, providing a unified framework compared to existing discretization-based or problem-specific neural solvers. MINPO was applied to a nonlinear Volterra IDE, reducing the memory operator error by approximately 33\% compared to A-PIKAN and 53\% compared to A-PINN. In the inverse setting, it achieved an average 60\% lower memory error than A-PIKAN and an order-of-magnitude improvement over A-PINN. In another experiment, MINPO was applied to three-dimensional nonlocal IDEs and reduced the memory operator error by nearly 98\% compared to A-PIKAN, corresponding to almost one order of magnitude improvement. Moreover, the proposed MINPO method demonstrated substantial accuracy gains and speedups over a finite-difference scheme when solving a three-dimensional nonlocal IDE. In time-fractional transport problems, MINPO maintained approximately 72\% lower memory error than fPINN and fPIKAN, showing robust performance even at high fractional orders.

\color{black}
\section{Acknowledgements}
S.A.F. acknowledges the support by the U.S. National Science Foundation under the Collaborations in Artificial Intelligence and Geosciences (CAIG, Award No. DE-2530611). 

\section{Code and Data Availability}
Codes and data available on our GitHub page.

\section{Conflict of Interest}
The authors declare no conflict of interests.

\def\mybibdoicolor{\color{black}}
\newcommand*{\doi}[1]{\href{\detokenize{#1}} {\raggedright\mybibdoicolor{DOI: \detokenize{#1}}}}

\bibliographystyle{elsarticle-num}
\bibliography{references_Memory.bib}

\newpage
\appendix
\setcounter{figure}{0}

\section{Preliminaries on PINN and PIKAN}\label{app:preliminary}

\subsection{Physics-Informed Neural Networks (PINN)} \label{Sec.PINN}

Physics-Informed Neural Networks (PINNs) approximate the solution of PDEs by embedding physical laws directly into the training process of a neural network~\cite{raissi2019physics}. Specifically, given a governing equation of the form, 
\begin{equation}
\mathcal{N}[u](\boldsymbol{x}, t) = 0, 
\qquad (\boldsymbol{x},t)\in\Omega\times(0,T),
\label{eq:pinn-pde}
\end{equation}
the goal is to learn a neural-network surrogate \( u(\boldsymbol{\xi};\boldsymbol{\theta}_{\text{MLP}}) \) that satisfies the operator \(\mathcal{N}[\cdot]\) in a soft, physics-informed manner.

The solution field is represented using a multilayer perceptron (MLP),
\begin{equation}
u(\boldsymbol{\xi}; \boldsymbol{\theta}_{\text{MLP}}) 
= \mathrm{MLP}(\boldsymbol{\xi}; \boldsymbol{\theta}_{\text{MLP}}),
\end{equation}
where \(\boldsymbol{\xi}=(\boldsymbol{x},t)\) and 
\(\boldsymbol{\theta}_{\text{MLP}}=\{W^{(\ell)},b^{(\ell)}\}_{\ell=1}^{L}\) denote the trainable parameters across \(L\) layers.

Each layer applies an affine transformation followed by a nonlinear activation,
\begin{equation}
h^{(\ell)} = \sigma\!\left(W^{(\ell)} h^{(\ell-1)} + b^{(\ell)}\right),
\qquad h^{(0)} = \boldsymbol{\xi},
\end{equation}
and produces the output,
\begin{equation}
h^{(L)} = u(\boldsymbol{\xi}; \boldsymbol{\theta}_{\text{MLP}}).
\end{equation}

The smooth differentiability of the MLP enables analytical computation of spatial and temporal derivatives via automatic differentiation, allowing the operator \(\mathcal{N}[u]\) in Equation~\eqref{eq:pinn-pde} to be evaluated directly within the network's computational graph.

\subsection{Physics-Informed Chebyshev-based Kolmogorov–Arnold Networks (cPIKANs)} \label{Sec.PIcKAN}

Physics-Informed Kolmogorov–Arnold Networks (PIKANs) extend the PINN framework by replacing the MLP with a Kolmogorov–Arnold Network (KAN)~\cite{li2024kolmogorov} based on the Kolmogorov–Arnold representation theorem ~\cite{kolmogorov1961representation}. In this work, we employ a Chebyshev-based variant~\cite{ss2024chebyshev}, termed cPIKAN, to represent the solution field,
\begin{equation}
u(\boldsymbol{\xi}; \boldsymbol{\theta}_{\text{cKAN}}) 
= \mathrm{cKAN}(\boldsymbol{\xi}; \boldsymbol{\theta}_{\text{cKAN}}),
\end{equation}
where \(\boldsymbol{\xi}=(\boldsymbol{x},t)\) and 
\(\boldsymbol{\theta}_{\text{cKAN}}\) denotes all trainable parameters of the cKAN architecture. 
As in the PINN formulation, the cPIKAN encodes the governing operator \(\mathcal{N}[\cdot]\), boundary conditions, initial conditions, and optional measurements directly into the loss function. 

Following the Kolmogorov–Arnold representation theorem, the cKAN architecture constructs multivariate functions through compositions of univariate transformations. Accordingly, we adopt the layer-wise mapping,
\begin{equation}\label{Eq:cKAN}
u(\boldsymbol{\xi};\boldsymbol{\theta}_{\text{cKAN}})
= (\boldsymbol{\Phi}_L \circ \tanh \circ \boldsymbol{\Phi}_{L-1} 
\circ \cdots \circ \tanh \circ \boldsymbol{\Phi}_1 \circ \tanh)(\boldsymbol{\xi}),
\end{equation}
where each layer \(\boldsymbol{\Phi}_l\) contains a matrix of univariate functions \(\phi_{l,i,j}\). Each \(\phi_{l,i,j}\) is expressed using a Chebyshev polynomial basis:
\begin{equation}
\phi_{l,i,j}(\xi)=\sum_{n=0}^{k} c_{l,i,j}^{(n)}\,T_n(\xi),
\qquad \xi\in[-1,1],
\end{equation}
with \(T_n(\boldsymbol{\xi})\) denoting the Chebyshev polynomials of the first kind,
\begin{equation}
T_0(\boldsymbol{\xi})=1,\qquad 
T_1(\boldsymbol{\xi})=\boldsymbol{\xi},\qquad
T_n(\boldsymbol{\xi})=2\boldsymbol{\xi} T_{n-1}(\boldsymbol{\xi})-T_{n-2}(\boldsymbol{\xi}),\quad n\ge 2.
\end{equation}

To ensure numerical stability during training and to respect the domain of Chebyshev polynomials, we apply a \(\tanh\) activation after each layer to constrain intermediate values within \([-1, 1]\).

The use of Chebyshev polynomials in the cPIKAN architecture offers several advantages: (i) Compact Representation: The spectral properties of Chebyshev polynomials allow for highly accurate function approximation with relatively few basis terms; (ii) Parameter Efficiency: The number of trainable parameters scales as \(O(N_l N_n^2 k)\), where \(N_l\) is the number of layers, \(N_n\) the number of neurons per layer, and \(k\) the polynomial degree; and (iii) Stability: Recursive definitions of Chebyshev polynomials are numerically stable, particularly near the domain boundaries, and mitigate issues that arise with trigonometric implementations during automatic differentiation.

Both the PINN and cPIKAN formulations are trained in an analogous manner. Let \(\mathcal{T}_{\text{res}} = \{\boldsymbol{\xi}^{\text{res}}_i\}_{i=1}^{N_{\text{res}}}\) be a set of interior collocation points in \(\Omega \times (0,T)\), and let \(\mathcal{T}_{\text{data}} = \mathcal{T}_{\text{init}} \cup \mathcal{T}_{\text{bc}} \cup \mathcal{T}_{\text{meas}}\) denote the union of initial, boundary, and optional measurement data. The total loss function is defined as,
\begin{equation}
\mathcal{L}(\boldsymbol{\theta}_{\text{Net}}) = \lambda_{\text{res}} \mathcal{L}_{\text{res}}(\mathcal{T}_{\text{res}}; \boldsymbol{\theta}_{\text{Net}}) + \lambda_{\text{data}} \mathcal{L}_{\text{data}}(\mathcal{T}_{\text{data}}; \boldsymbol{\theta}_{\text{Net}}),
\end{equation}
where
\begin{align}
\mathcal{L}_{\text{res}} &= \frac{1}{N_{\text{res}}} \sum_{i=1}^{N_{\text{res}}} \left| \mathcal{N}\left[u(\boldsymbol{\xi}_i^{\text{res}}; \boldsymbol{\theta}_{\text{Net}})\right] \right|^2, \\
\mathcal{L}_{\text{data}} &= \frac{1}{N_{\text{data}}} \sum_{j=1}^{N_{\text{data}}} \left| u(\boldsymbol{\xi}_j^{\text{data}}; \boldsymbol{\theta}_{\text{Net}}) - u_j^{\text{data}} \right|^2.
\end{align}
The coefficients \(\lambda_{\text{res}}\) and \(\lambda_{\text{data}}\) control the relative weight of the physics-based and data-driven constraints. \(\boldsymbol{\theta}_{\text{Net}}\) denotes the collection of all trainable parameters, corresponding to either the MLP or the cKAN.

\subsection{Scaled-cPIKAN} \label{Sec.ScaledcPIKAN}

cPIKANs are highly effective for approximating solutions to PDEs on compact domains. However, when applied to problems defined over extended spatial domains or involving high-frequency solution components, standard cPIKAN implementations may suffer from numerical instabilities. Although introducing \(\tanh\) nonlinearities between layers helps constrain intermediate activations within \([-1,1]\), such regularization alone is often insufficient to ensure stable and accurate training, particularly in stiff or oscillatory regimes. To overcome these challenges, the Scaled-cPIKAN framework was developed~\cite{mostajeran2025scaled} that introduces a spatial scaling strategy in which the original PDE is reformulated over a standardized domain \(\tilde{\Omega} = [-1,1]^d\). All components of the PDE, including differential operators, source terms, and boundary or initial conditions, are consistently transformed into this scaled coordinate system. Correspondingly, all training data (residual points, boundary data, and measurements) are mapped into the scaled domain to ensure physical and numerical consistency. This rescaling approach offers two principal advantages. First, it aligns the spatial inputs with the natural domain of Chebyshev polynomials, thereby preserving their orthogonality and improving numerical conditioning. Second, it enhances the stability of the training process, particularly in scenarios with sharp gradients or multiscale structures, by mitigating the distortion of derivatives at domain boundaries. The Scaled-cPIKAN formulation retains the architectural and spectral advantages of Chebyshev-based networks while significantly broadening their applicability to large domains and challenging PDEs. By reformulating the physics in a normalized domain, this approach improves robustness without compromising the fidelity of the learned solutions.

\section{A-PIKAN Formulation}\label{A-PIKANN}

\begin{figure}[h]
    \centering
    \includegraphics[width=0.99\linewidth]{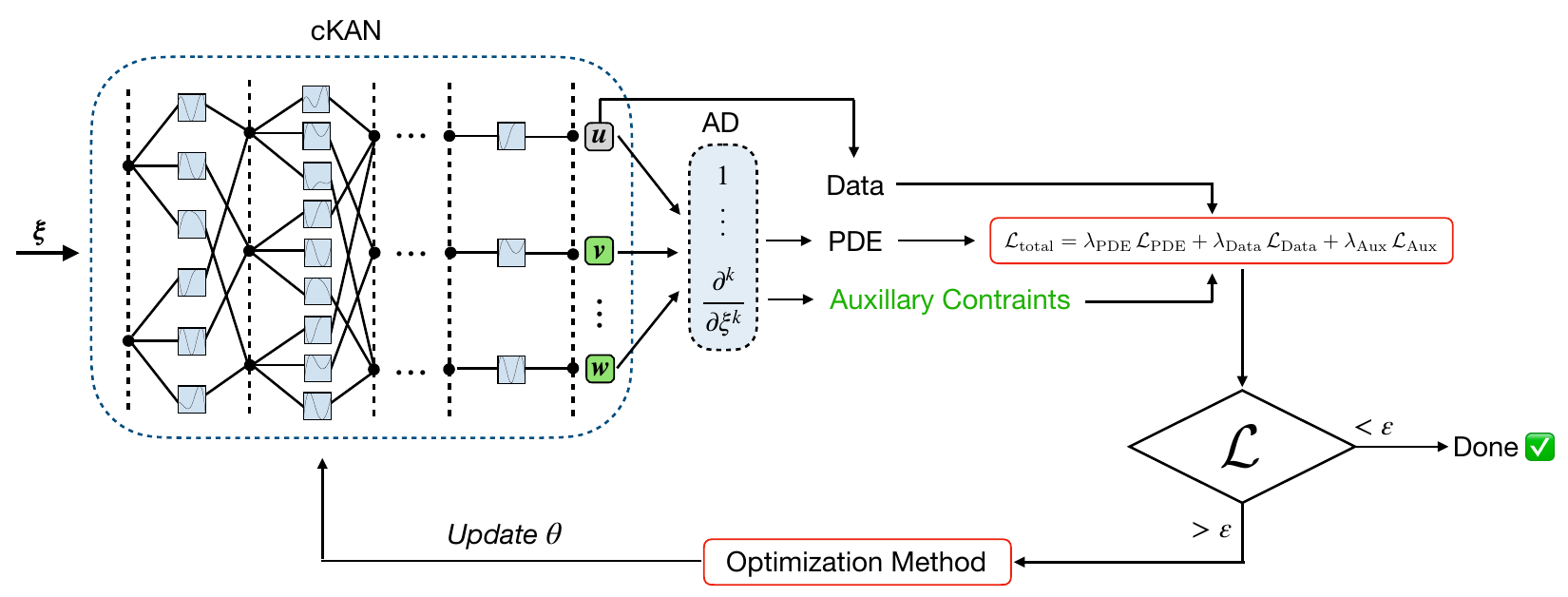}
    \caption{A schematic architecture of the Chebyshev-based Auxiliary PIKAN (A-PIKAN) framework for solving IDEs. The cKAN network takes the input coordinates $\boldsymbol{\xi}$ and outputs both the primary field $u$ and a set of auxiliary variables $[v,...,w]$ that represent the integral terms. These auxiliary outputs flow through the computational graph together with $u$, where automatic differentiation (AD) computes all required spatial and temporal derivatives. The PDE residual and the auxiliary-evolution residual, combined with initial and boundary data, form the total loss used during training.}
    \label{fig:A-PIKAN}
\end{figure}

While standard PIKANs enforce differential constraints via automatic differentiation, they require explicit discretization of nonlocal operators when applied to IDEs. Such quadrature-based discretization introduces truncation errors and significantly increases computational cost, especially for higher-order kernels or dense collocation sampling. To overcome these limitations, we extend the auxiliary-variable concept of the A-PINN framework ~\cite{yuan2022pinn} to the Chebyshev-based formulation, resulting in the \textit{Chebyshev-based Auxiliary PIKAN (A-PIKAN)}. The overall architecture of the proposed A-PIKAN framework is illustrated in Fig.~\ref{fig:A-PIKAN}, which highlights the auxiliary-variable outputs and Chebyshev-cKAN.

Let a nonlinear IDE be expressed as,
\begin{equation}
\mathcal{N}[u](\boldsymbol{\xi}) = f(\boldsymbol{\xi}) 
+ \lambda \!\!\int_{g(\boldsymbol{\xi})}^{h(\boldsymbol{\xi})} 
K(\boldsymbol{\xi},\tau)\,u(\tau)\,d\tau,
\end{equation}
where $\mathcal{N}[\cdot]$ is a differential operator, $K(\boldsymbol{\xi},\tau)$ is a known kernel, and $\lambda$ is a coupling parameter. We define an auxiliary output $v(\boldsymbol{\xi})$ to approximate the integral term,
\begin{equation}
v(\boldsymbol{\xi}) = \!\!\int_{g(\boldsymbol{\xi})}^{h(\boldsymbol{\xi})} 
K(\boldsymbol{\xi},\tau)\,u(\tau)\,d\tau,
\end{equation}
and reformulate the IDE equivalently as a coupled system,
\begin{equation}
\begin{cases}
\mathcal{N}[u](\boldsymbol{\xi}) = f(\boldsymbol{\xi}) + \lambda\,v(\boldsymbol{\xi}),\\[3pt]
\dfrac{\partial v(\boldsymbol{\xi})}{\partial t} = \widetilde{K}(t)\,u(\boldsymbol{\xi}),\\[3pt]
u(\boldsymbol{x},0)=u_0(\boldsymbol{x}),\quad v(\boldsymbol{x},0)=0.
\end{cases}
\label{Eq:AuxSystem}
\end{equation}
As shown in Fig.~\ref{fig:A-PIKAN}, this auxiliary-variable formulation is encoded as an additional output that evolves according to the Eq.~\eqref{Eq:AuxSystem}. 
The nonlocal operator is thus replaced by a differentiable constraint that can be enforced via automatic differentiation of the network outputs, eliminating the need for explicit quadrature. The Chebyshev polynomial basis provides improved spectral accuracy and parameter efficiency, enabling more compact networks.

Following the A-PINN framework~\cite{yuan2022pinn}, the A-PIKAN loss function consists of three components:
(i) the residual of the governing equation,
(ii) the auxiliary-variable constraint, and
(iii) the initial and/or boundary conditions. 
Let the predicted outputs of the A-PIKAN be $
u_{\text{pred}}(\boldsymbol{\xi}_i; \boldsymbol{\theta}_{\text{A-PIKAN}})$ and 
$v_{\text{pred}}(\boldsymbol{\xi}_i; \boldsymbol{\theta}_{\text{A-PIKAN}})$
at collocation points $\boldsymbol{\xi}_i = (\boldsymbol{x}_i,t_i)$.
The auxiliary constraint~\eqref{Eq:AuxSystem} is imposed through automatic differentiation of $v_{\text{pred}}$ with respect to time,
\begin{equation}
\mathcal{L}_{\text{Aux}} 
= \frac{1}{N_{\text{PDE}}}
\sum_{i=1}^{N_{\text{res}}}
\left|
\dfrac{\partial v_{\text{pred}}(\boldsymbol{\xi}_i;\boldsymbol{\theta}_{\text{A-PIKAN}})}{\partial t}
- K(t_i)\,u_{\text{pred}}(\boldsymbol{\xi}_i;\boldsymbol{\theta}_{\text{A-PIKAN}})
\right|^2.
\label{Eq:Laux}
\end{equation}

This equation corresponds to the auxiliary branch in Fig.~\ref{fig:A-PIKAN}, where the time derivative of $v_{\text{pred}}$ is enforced through automatic differentiation.

The residual of the governing IDE is computed as,
\begin{equation}
\mathcal{L}_{\text{PDE}}
= \frac{1}{N_{\text{PDE}}}
\sum_{i=1}^{N_{\text{PDE}}}
\left|
\mathcal{N}\!\left[u_{\text{pred}}(\boldsymbol{\xi}_i;\boldsymbol{\theta}_{\text{A-PIKAN}})\right]
- f(\boldsymbol{\xi}_i)
- \lambda\,v_{\text{pred}}(\boldsymbol{\xi}_i;\boldsymbol{\theta}_{\text{A-PIKAN}})
\right|^2,
\label{Eq:Lres}
\end{equation}
and the loss corresponding to initial and boundary conditions is defined as,
\begin{equation}
\mathcal{L}_{\text{data}}
= \frac{1}{N_{\text{data}}}
\sum_{j=1}^{N_{\text{data}}}
\Big(
\big|u_{\text{pred}}(\boldsymbol{\xi}_j^{\text{data}};\boldsymbol{\theta}_{\text{cKAN}})
- u_j^{\text{data}}\big|^2
+ 
\big|v_{\text{pred}}(\boldsymbol{\xi}_j^{\text{data}};\boldsymbol{\theta}_{\text{cKAN}})
- v_j^{\text{data}}\big|^2
\Big),
\label{Eq:Ldata}
\end{equation}
where $(u_j^{\text{data}},v_j^{\text{data}})$ denote the prescribed conditions.

The total A-PIKAN loss is then given as the weighted sum of these three components,
\begin{equation}
\mathcal{L}_{\text{total}}
= \lambda_{\text{PDE}}\,\mathcal{L}_{\text{PDE}}
+ \lambda_{\text{Data}}\,\mathcal{L}_{\text{Data}} 
+ \lambda_{\text{Aux}}\,\mathcal{L}_{\text{Aux}},
\label{Eq:Ltotal}
\end{equation}
where $\lambda_{\text{res}}, \lambda_{\text{aux}},$ and $\lambda_{\text{data}}$ are positive weighting coefficients controlling the relative importance of each term.

\section{fPIKAN Formulation}\label{fPIKANN}

Within the unified integro–differential framework introduced in Eq.~\eqref{eq:general-IDE}, fractional partial differential equations (fPDEs), correspond to the special case in which all nonlocal effects arise from temporal memory encoded through the fractional evolution operator \(\mathcal{T}_\alpha\). In this setting, the governing equation takes the form,
\begin{equation}
\mathcal{T}_\alpha[u](\boldsymbol{\xi})
=
\mathcal{N}[u](\boldsymbol{\xi})
+
S(\boldsymbol{\xi}),
\label{Eq:fPDE_FPIKAN}
\end{equation}
where operator \(\mathcal{T}_\alpha\) combines classical and fractional time dynamics as defined in Eq.~\eqref{eq:T-operator}. When the classical contribution is removed \( (\lambda_1=0)\), the equation reduces to a pure time-fractional PDE whose temporal nonlocality is fully described by the Caputo derivative \({}^{C}D_t^{\alpha}\), given in Eq.~\eqref{Eq:CaputoDeriv_1}. This operator accounts for the history-dependent behavior through a weighted integral over all past time, reflecting the intrinsic memory effects present in the system.

\begin{lemma}\label{lem:CaputoIdentity}
Let \(\mathcal{I}^{\alpha}\) denote the fractional integral of order \(\alpha\in(0,1)\) with base point \(0\), defined by~\cite{kilbas2006theory},
\begin{equation}\label{Eq:FractionalIntegral}
\mathcal{I}^{\alpha} h(t)
=\frac{1}{\Gamma(\alpha)}
\int_{0}^{t} (t-\tau)^{\alpha-1} h(\tau)\, d\tau .
\end{equation}

Then the Caputo fractional derivative satisfies the identity~\cite{kilbas2006theory},
\begin{equation}
\mathcal{I}^{\alpha}\bigl({}^{C}D_t^{\alpha} h(t)\bigr)
=h(t)-h(0),
\end{equation}
for all functions \(h\) that are absolutely continuous on \([0,t]\).
This relation expresses the Caputo derivative as the left-inverse of the fractional integral operator, up to the initial value \(h(0)\), and highlights the intrinsic coupling between fractional differentiation, memory accumulation, and initialization.
\end{lemma}

\begin{figure}[h]
    \centering
    \includegraphics[width=0.99\linewidth]{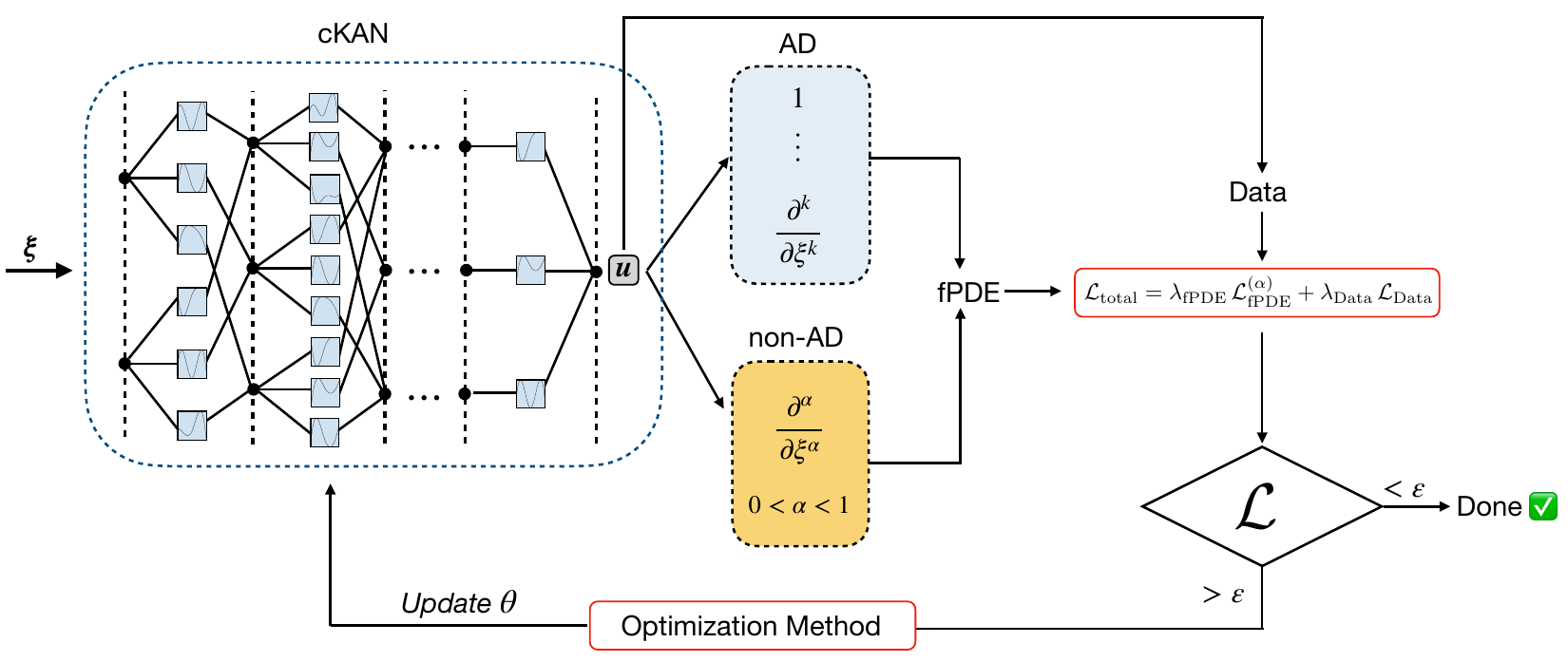}
    \caption{A schematic architecture for the fractional cPIKAN (fPIKAN) framework. The cKAN network takes the input coordinates $\boldsymbol{\xi}$ and outputs the primary field $u$. Integer-order derivatives of the output are evaluated through automatic differentiation (AD), while fractional derivatives $(0 < \alpha < 1)$ are discretized (non-AD) using schemes such as the L1 method for the Caputo time derivative. The AD-based derivatives and the discretized fractional components (non-AD) together define the fractional IDE residual. This residual, combined with initial and boundary data or sparse measurements, forms the total loss used during training.}
    \label{fig:fPIKAN}
\end{figure}

Following the methodology of fractional Physics-Informed Neural Networks (fPINNs)~\cite{pang2019fpinns}, we extend the framework to a Chebyshev-based formulation architecture termed the \textit{fractional cPIKAN (fPIKAN)}. The fPIKAN, as shown in Fig.~\ref{fig:fPIKAN}, inherits the fractional-physics formulation of fPINN but replaces the conventional MLP with the cKAN, enabling higher spectral accuracy and better parameter efficiency when modeling nonlocal fractional operators. The architecture of the proposed fPIKAN framework is illustrated in Fig.~\ref{fig:fPIKAN}.

In  Eq.~\eqref{Eq:fPDE_FPIKAN},  \(\mathcal{N}[u]\) contains all local terms that can be evaluated directly using automatic differentiation (AD), while \(\mathcal{T}_\alpha[u]\) represents the memory contributions that require numerical quadrature or specialized approximation. As illustrated in Fig.~\ref{fig:fPIKAN}, these AD-based and non-AD components follow separate computational paths within the fPIKAN architecture.
We approximate the Caputo derivative using the classical L1 finite-difference scheme~\cite{gao2014new}.
Given a temporal grid \({t_k}_{k=0}^{N_t}\) with uniform step size $\Delta t = t/N_t$, 
the Caputo derivative of order \(\alpha \in (0,1)\) is approximated as~\cite{gao2014new},
\begin{equation}\label{Eq:L1}
\frac{\partial^{\alpha} u(\boldsymbol{x},t_n)}{\partial t^{\alpha}}
\;\approx\;
\frac{(\Delta t)^{-\alpha}}{\Gamma(2-\alpha)}
\left[
c_0\,u(\boldsymbol{x},t_n)
+
\sum_{k=1}^{n-1}
\big( c_{n-k} - c_{n-k-1} \big)\,
u(\boldsymbol{x},t_k) \, +\, c_{n-1}\,u(\boldsymbol{x},0)
\right],
\end{equation}
where the coefficients,
\begin{equation}
    c_\ell = (\ell+1)^{1-\alpha} - \ell^{1-\alpha}, \qquad \ell = 0,\dots,n-1,
\end{equation}
encode the fractional-order memory weights.
Here \(N_t\) controls the temporal resolution of the discretization.

Let $u_{\text{pred}}(\boldsymbol{\xi}; \boldsymbol{\theta}_{\text{fPIKAN}})$ denote the network-predicted solution obtained from the fPIKAN model parameterized by $\boldsymbol{\theta}_{\text{fPIKAN}}$.  
The residual of the fPDE~\eqref{Eq:fPDE_FPIKAN} at a collocation point $\boldsymbol{\xi}_i$ is defined as,
\begin{equation}
\mathcal{R}^{(\alpha)}(\boldsymbol{\xi}_i)
= \mathcal{T}_\alpha\! \left[u_{\text{pred}}\right](\boldsymbol{\xi}_i; \boldsymbol{\theta}_{\text{fPIKAN}})
- \mathcal{N}\! \left[u_{\text{pred}}\right](\boldsymbol{\xi}_i; \boldsymbol{\theta}_{\text{fPIKAN}})
- S(\boldsymbol{\xi}_i),
\end{equation}
hence the physics-based residual loss is then evaluated as,
\begin{equation}
\mathcal{L}_{\text{fPDE}}^{(\alpha)} =
\frac{1}{N_{\text{res}}}
\sum_{i=1}^{N_{\text{res}}}
\big|\mathcal{R}^{(\alpha)}(\boldsymbol{\xi}_i)\big|^2,
\end{equation}
while the data loss enforces boundary and initial conditions as,
\begin{equation}
\mathcal{L}_{\text{data}} =
\frac{1}{N_{\text{data}}}
\sum_{j=1}^{N_{\text{data}}}
\big|u_{\text{pred}}(\boldsymbol{\xi}_j^{\text{data}}; \boldsymbol{\theta}_{\text{fPIKAN}})
- u_j^{\text{data}}\big|^2.
\end{equation}

The total loss used for training the fPIKAN model is defined as,
\begin{equation}
\mathcal{L}_{\text{total}}
= \lambda_{\text{fPDE}}\,\mathcal{L}_{\text{fPDE}}^{(\alpha)}
+ \lambda_{\text{Data}}\,\mathcal{L}_{\text{Data}},
\label{Eq:Ltotal_FPIKAN}
\end{equation}
where $\lambda_{\text{fPDE}}$ and $\lambda_{\text{data}}$ are tunable weights balancing the physics and data constraints.

\section{Solution Field  Reconstruction  from  Memory Operator}\label{app:inv_M}
\subsection{Reconstruction Procedure for Experiment I}\label{app:inv_M-I}
To illustrate the reconstruction procedure in Eq.~\eqref{Eq:VolterraInverse}, we start from the definition of the Volterra-type operator, 
\begin{equation}
    \mathcal{M}[u](t)=\int_0^{t} e^{(\tau-t)}\, u(\tau),d\tau .
\end{equation}

Differentiating this expression with respect to \(t\) using Leibniz’ rule gives~\cite{osler1972integral},
\begin{equation}
    \frac{d}{dt}\mathcal{M}[u](t)=u(t)-\mathcal{M}[u](t),
\end{equation}
as the boundary term contributes \(u(t)\) and the derivative of the kernel \(e^{(\tau-t)}\) equals \(-e^{(\tau-t)}\).
Rearranging this identity yields the explicit inverse mapping of the Volterra operator,
\begin{equation}
    u(t)=\frac{d}{dt}\mathcal{M}[u](t)+\mathcal{M}[u](t).
\end{equation}

By substituting the learned nonlocal model \(\mathcal{M}_{\boldsymbol{\theta}}\) in place of \(\mathcal{M}[u]\), we obtain the practical reconstruction formula used in the implementation,
\begin{equation}
    u_{\boldsymbol{\Theta}}(t)=\partial_t \mathcal{M}_{\boldsymbol{\theta}}(t)+\mathcal{M}_{\boldsymbol{\theta}}(t),
\end{equation}
where the spatial derivative \(\partial_t \mathcal{M}_{\boldsymbol{\theta}}\) is computed efficiently via automatic differentiation.

\subsection{Reconstruction Procedure for Experiment II}\label{app:inv_M-3D}

To illustrate the reconstruction procedure in Eq.~\eqref{Eq:3DInverse}, we start from the definition of the three-dimensional Volterra-type operator,
\begin{equation}
\mathcal{M}[u](\boldsymbol{\xi})
= \int_{0}^{x_2}\,\int_{0}^{x_1}\,\int_{0}^{t}
e^{(\tau - t)}\,
u(y_1,y_2,\tau)\,
d\tau\, dy_1\, dy_2,
\qquad
\boldsymbol{\xi}=(x_1,x_2,t)\in\Omega.
\end{equation}

Because the integration limits depend on the spatial coordinates and the time variable, repeated application of Leibniz’ rule~\cite{osler1972integral} gives the identity,
\begin{equation}\label{Eq:3DLeibniz}
\frac{\partial^3}{\partial t\,\partial x_1\,\partial x_2}\,
\mathcal{M}[u](\boldsymbol{\xi})
=e^0\,u(x_1,x_2,t)
-
\frac{\partial^2}{\partial x_1\,\partial x_2}\,
\mathcal{M}[u](\boldsymbol{\xi}),
\end{equation}
where, the first term on the right-hand side corresponds to differentiating the integral through all three variable upper bounds, and the remaining term arises from the derivative of the kernel \(e^{(\tau - t)}\) with respect to \(t\).
Since \(e^0=1\), Eq.~\eqref{Eq:3DLeibniz} rearranges to,
\begin{equation}
u(\boldsymbol{\xi})
= \frac{\partial^3\mathcal{M}[u]}{\partial t\,\partial x_1\,\partial x_2}
+
\frac{\partial^2\mathcal{M}[u]}{\partial x_1\,\partial x_2}.
\end{equation}

By replacing \(\mathcal{M}[u]\) with the learned operator \(\mathcal{M}_{\boldsymbol{\theta}}\), we obtain the practical reconstruction formula used in the implementation,
\begin{equation}
u_{\boldsymbol{\Theta}}(\boldsymbol{\xi})
= \frac{\partial^3 \mathcal{M}_{\boldsymbol{\theta}}}{\partial t\,\partial x_1\,\partial x_2}
+
\frac{\partial^2 \mathcal{M}_{\boldsymbol{\theta}}}{\partial x_1\,\partial x_2},
\qquad
\boldsymbol{\xi}\in\Omega.
\end{equation}

All mixed partial derivatives are evaluated using automatic differentiation, ensuring a consistent and fully continuous reconstruction of the unknown field.

\end{document}